\documentclass{article} 
\usepackage{iclr2026_conference,times}

\usepackage{amsmath,amsfonts,bm}

\def\eqref#1{equation~\ref{#1}}

\def\1{\bm{1}}

\DeclareMathAlphabet{\mathsfit}{\encodingdefault}{\sfdefault}{m}{sl}
\SetMathAlphabet{\mathsfit}{bold}{\encodingdefault}{\sfdefault}{bx}{n}

\usepackage{hyperref}
\usepackage{adjustbox}
\usepackage{url}
\usepackage{graphicx}
\usepackage{booktabs}
\usepackage{transparent}
\usepackage{colortbl}
\usepackage{nicematrix}
\usepackage{silence}
\usepackage{csquotes}
\ActivateWarningFilters[pdftoc]

\newcommand{\NA}{---}

\title{More Thought, Less Accuracy? On the Dual Nature of Reasoning in Vision-Language Models}

\author{%
Xinyu Tian\textsuperscript{\dag},
Shu Zou\textsuperscript{\dag \S},
Zhaoyuan Yang\textsuperscript{*},
Mengqi He\textsuperscript{\dag},
Fabian Waschkowski\textsuperscript{\ddag\S},\\
\textbf{Lukas Wesemann\textsuperscript{\ddag\S},} 
\textbf{Peter Tu\textsuperscript{*},}
\textbf{Jing Zhang\textsuperscript{\dag},} \\
\textsuperscript{\dag}Australian National University \quad
\textsuperscript{\ddag}University of Melbourne \quad
\textsuperscript{*}GE Research \quad 
\textsuperscript{\S}Maincode \\ 
\texttt{\small{firstname.lastname@anu.edu.au, firstname.lastname@ge.com}}
}

\iclrfinalcopy 
\begin{document}

\maketitle


\begin{abstract}
Reasoning has emerged as a pivotal capability in Large Language Models (LLMs). Through Reinforcement Learning (RL), typically Group Relative Policy Optimization (GRPO), these models are able to solve complex tasks such as mathematics and code generation. Building on these advances, recent research has sought to extend reasoning to Vision-Language Models (VLMs), yielding promising results across diverse visual tasks. Despite this progress, our study uncovers the dual nature of multimodal reasoning: while it substantially enhances logical inference and facilitates performance on challenging problems, longer reasoning length may gradually impair perceptual grounding, leading to recognition failures on otherwise basic visual questions. Through further analysis, we attribute this phenomenon to visual forgetting, wherein prolonged reasoning length causes models to disregard visual input. To address this, we propose {\sc Vision-Anchored Policy Optimization} (\texttt{VAPO}), a simple yet effective method that explicitly steers the reasoning process toward visually grounded trajectories. Our result model, \texttt{VAPO-Thinker-7B}, significantly strengthens the model's reliance on visual information and achieves new state-of-the-art results on various benchmarks.
\end{abstract}

\section*{1 \quad Introduction}

    Reasoning has long been recognized as an essential capability of Large Language Models (LLMs). Early approaches, such as chain-of-thought~\citep{wei2022chain, kojima2022large}, encourage models to produce step-by-step explanations before arriving at an answer, whereas more recent advances have introduced aha moments~\citep{guo2025deepseek, muennighoff2025s1}, characterized by cognitive behaviors such as self-reflection and verification. These developments have enabled models to address increasingly complex problems, \eg, mathematics and coding. Consequently, the community has recently sought to integrate reasoning into Vision-Language Models (VLMs). By leveraging Reinforcement Learning (RL) such as Group Relative Policy Optimization (GRPO)~\citep{shao2024deepseekmath}, researchers have trained models to think while tackling challenging multimodal tasks, \eg, geometry or navigation~\citep{yang2025r1, huang2025vision}. Remarkably, this integration has also demonstrated promising results on traditional visual tasks, \eg, classification and detection~\citep{liu2025visual, shen2025vlm}, while yielding stronger generalization~\citep{chu2025sft}.

Building on these successful practices, reasoning has emerged as a powerful and seemingly universal strategy for addressing a wide range of visual tasks. However, its limitations remain largely unexplored. Early doubts are raised by \citet{li2025think}, which find that in certain scenarios, models trained with explicit reasoning result in only marginal gains compared to direct answering. More recently, \citet{xia2025visionary} observe a counter-intuitive trend: unlike in text domains, RL in multimodal settings often exhibits a collapse. As accuracy improves during training, the reasoning process gradually shortens, suggesting that models tend to favor brief, less structured responses. These findings lead us to ask: \textit{Is reasoning truly a consistent performance booster for vision-language models?}

\begin{figure*}[t!] 
    \centering
    \includegraphics[width=1\linewidth]{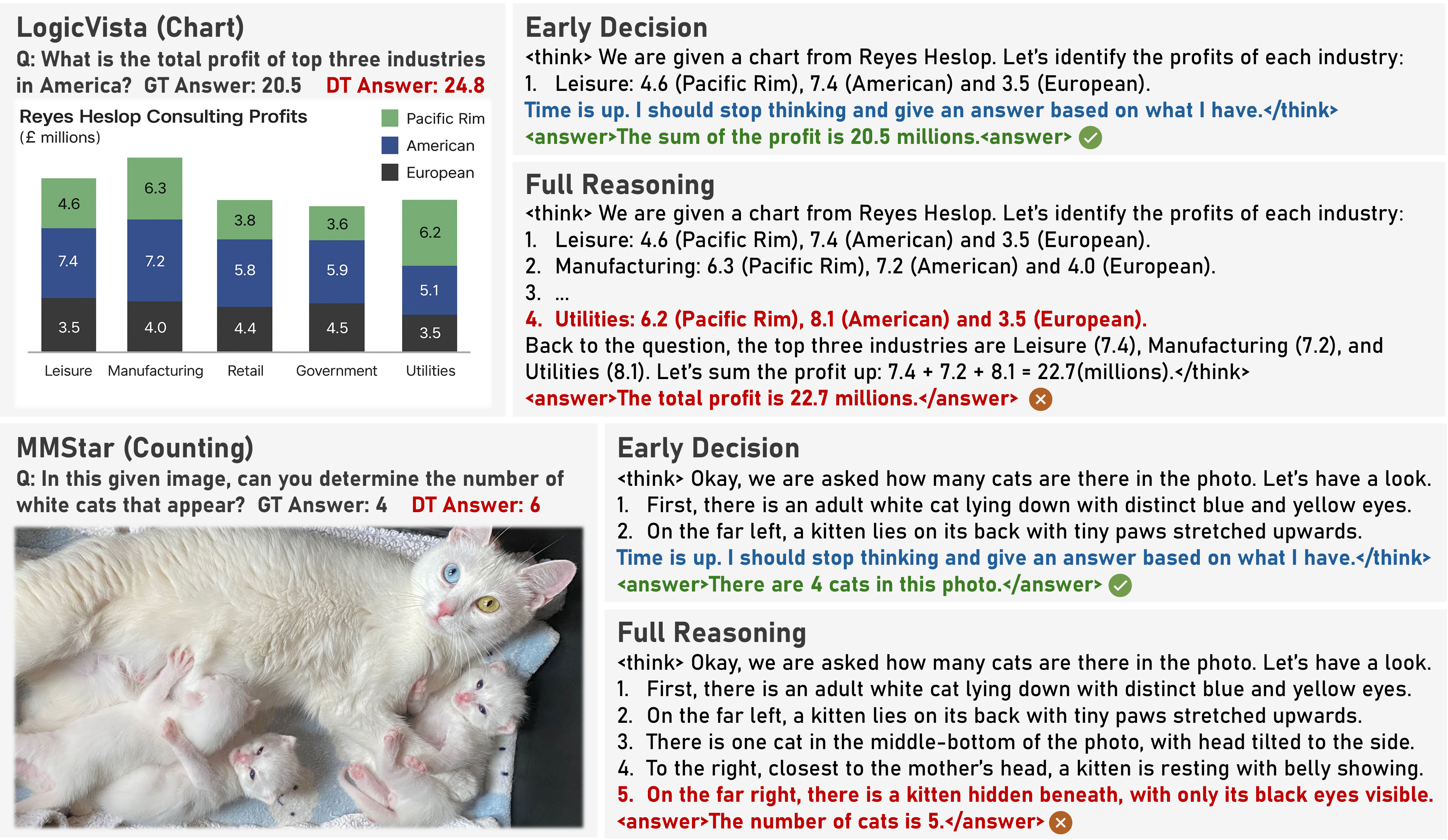}
    \vspace{-6mm}
    \caption{The examples where more reasoning yields less accuracy. We select Vision-R1~\citep{huang2025vision} as a representitive multimodal reasoning model and evaluate on established benchmarks. Given the ground truth (GT) answer, we examine three settings, \ie, direct (DT) answering, full reasoning, and early decision. Correct and incorrect responses are highlighted in \textcolor[HTML]{3B7D23}{green} and \textcolor[HTML]{C00000}{red}, respectively, while designed prompts for early decision are indicated in \textcolor[HTML]{215F9A}{blue}.
    }
\label{fig:motivation}
\vspace{-6mm}
\end{figure*}


Motivated by this question, we perform a zoom-in analysis of how the reasoning process impacts accuracy of existing VLMs. Beyond the standard settings of direct answering and full-step reasoning, we introduce an early decision mode, where the model is prompted to halt reasoning at selected logical breakpoints and produce an immediate answer. This setup allows us to measure the contribution of each reasoning segment to the final outcome. Interestingly, our finding reveals that while early-stage reasoning may offer substantial gains, its advantages tend to saturate and may even reverse in later steps. To understand this phenomenon, we conduct an error analysis of the reasoning content in failure cases and find that perception errors, where the model incorrectly interprets visual details in the image, constitute the dominant category, accounting for over $50\%$. Moreover, the majority of these errors can be rectified with a shorter reasoning via early decision, indicating the model is initially capable of producing correct answers but is ultimately misled by extended reasoning. For instance, as shown in Fig.~\ref{fig:motivation}, the early stage of reasoning paths, \ie, early decision, enables the model to arrive at correct answers, whereas prolonged thoughts, \ie, full reasoning, often lead to errors such as chart misreading or hallucination. This indicates the dual nature of reasoning: while it strengthens logical inference and facilitates complex problem-solving, longer reasoning length may weaken perceptual grounding, leading to susceptibility on otherwise straightforward visual questions.


This decline in perceptual capability constitutes a major bottleneck for multimodal reasoning. To further investigate its underlying causes, we examine how attention to visual tokens evolves throughout the reasoning process. Our analysis reveals a marked decrease in visual attention as reasoning progresses, suggesting the emergence of reasoning may inadvertently reduce the model's reliance on visual input, a phenomenon we refer to as visual forgetting. To validate this hypothesis, we propose two interventions: 1) visual replay, which reintroduces the input image to models at regular intervals during reasoning; 2) focus prompt, which prompts models to attend to the visual input at selected steps. Both methods are proven to alleviate performance degradation induced by reasoning, providing evidence that visual forgetting is a central factor constraining reasoning potentials.


While the two aforementioned remedies can partially alleviate visual forgetting, they incur substantial computational overhead during inference and fail to rectify the underlying behavioral deficiencies of existing reasoning models. To address this, we propose {\sc Vision-Anchored Policy Optimization} (\texttt{VAPO}), a simple yet effective policy gradient algorithm that explicitly guides the reasoning process toward a visually grounded trajectory. The key idea is to embed a sequence of visual anchors throughout the reasoning path. At each anchor, the model's perceptual capability is assessed by evaluating its responses to a set of primitive visual claims. Beyond standard outcome-based rewards such as accuracy and format, we introduce perception reward, which quantifies the model's overall perceptual grounding during reasoning by aggregating scores across all anchor points. Experimental results demonstrate that our result model, \texttt{VAPO-Thinker-7B}, significantly enhances the model's reliance on visual input during reasoning and achieves average gains of $2\sim4\%$ over strong baselines on established benchmarks, setting a new state of the art.

In summary, our contributions are as follows:
\begin{itemize}
    \item Despite the promise of multimodal reasoning, we identify a pronounced dual nature, where reasoning enhances logical inference capability at the expense of perceptual accuracy as the length grows, constituting a major bottleneck to the overall effectiveness.
    \item We further examine this side effect of reasoning and empirically demonstrate that it stems from visual forgetting, where prolonged reasoning length gradually reduces the model's reliance on visual input, leading to a substantial increase in perceptual failures.
    \item We propose \texttt{VAPO}, a straightforward yet effective policy gradient algorithm designed to strengthen the model's dependence on visual information during reasoning, and demonstrate its effectiveness through experimental validation.
\end{itemize}

\vspace{-1mm}
\section*{2 \quad Related Work}

\textbf{Reasoning in Large Language Models}. Conventional research has posited a trade-off between interpretability and accuracy~\citep{wang2020high, koh2020concept, yao2023training}. The advent of reasoning, however, has emerged as a notable exception to the paradigm. Early works demonstrate that prompting LLMs~\citep{wei2022chain, yao2023tree, zou2025unlocking, tian2024argue} or applying Supervised Fine-Tuning (SFT)~\citep{yang2025qwen3, cai2024internlm2} to encourage step-by-step explanations prior to answering could lead to substantial gains. These findings underscore the critical role of constructing large-scale, high-quality chain-of-thought trajectories. More recently, the emergence of RL-based methods such as GRPO~\citep{shao2024deepseekmath} has diminished the reliance on manually crafted reasoning paths. With only a limited number of examples, models are now capable of autonomously discovering optimal reasoning strategies and even exhibiting advanced behaviors such as self-reflection and verification~\citep{yu2025dapo, zheng2025group}. Consequently, Test-Time Scaling (TTS)~\citep{muennighoff2025s1, snell2024scaling, qu2025optimizing, yao2025simple} has become a standard practice in LLMs, driven by the belief that longer or more elaborate reasoning consistently yields better performance. However, our empirical analysis challenges the universality of this assumption, revealing that TTS does not necessarily generalize to multimodal settings, particularly regarding its implications for VLMs.

\textbf{Reasoning in Vision-Language Models}. Building on the success of reasoning in LLMs, a growing line of research has sought to extend this capability to VLMs~\citep{wang2025think, meng2025mm, vl-rethinker, he2025few, tian2025identifying, tian2025black, zou2025simlabel}. Two predominant approaches have emerged: one leverages limited high-quality trajectories for cold-start via SFT, followed by RL~\citep{peng2025skywork, yang2025look, tan2025reason, feng2025video}, while the other bypasses SFT entirely, employing RL alone to discover optimal reasoning paths~\citep{liu2025visual, wang2025perception}. Both approaches have yielded promising results on various visual tasks, particularly those requiring complex logical inference. However, recent findings indicate that, in certain scenarios, models trained with explicit reasoning offer little or no gains over direct answering~\citep{li2025think}. Moreover, training dynamics reveal that as accuracy increases, the length of generated reasoning tends to diminish, implying a preference for shorter and more concise responses~\citep{xia2025visionary}. These preliminary findings indicate that reasoning may not be a free lunch in the development of VLMs. To the best of our knowledge, our work presents the first systematic investigation into the double-edged nature of reasoning in VLMs, identifying visual forgetting as a key limitation and proposing effective solutions to mitigate this issue.

\textbf{Forgetting in Vision-Language Models}. In fact, visual forgetting has long been a well-known yet insufficiently addressed issue. Since the emergence of VLMs, researchers have identified a pronounced text bias in these LLM-backboned models, where responses are predominantly guided by textual input while visual cues are largely ignored~\citep{chen2024image, fu2024blink}. At the time, this modality imbalance is less apparent, as early VLMs typically generate short outputs and lack sophisticated reasoning capabilities. Consequently, most prior efforts focus on test-time remedies such as contrastive decoding~\citep{leng2024mitigating, wang2024mitigating} or attention reallocation~\citep{tu2025attention, gong2024damro} to circumvent visual forgetting. However, with the recent rapid progress in multimodal reasoning, VLMs have increasingly shifted from generating short answers to producing extended, step-by-step explanations. This shift has inadvertently amplified the forgetting issue, limiting the benefits of reasoning in multimodal contexts. To address this, we propose \texttt{VAPO}, a training-based method that explicitly reinforces the model's perceptual grounding, effectively alleviating visual forgetting, and achieving a new state of the art across established benchmarks.

\vspace{-1mm}
\section*{3 \quad The Dual Nature of Reasoning}
\label{sec:3}

\vspace{-1mm}
\subsection*{3.1 \quad No-Free-Lunch Dilemma: Logic vs. Perception}
\label{subsec:3.1}

\begin{figure*}[t!] 
    \centering
    
    \includegraphics[width=1\linewidth]{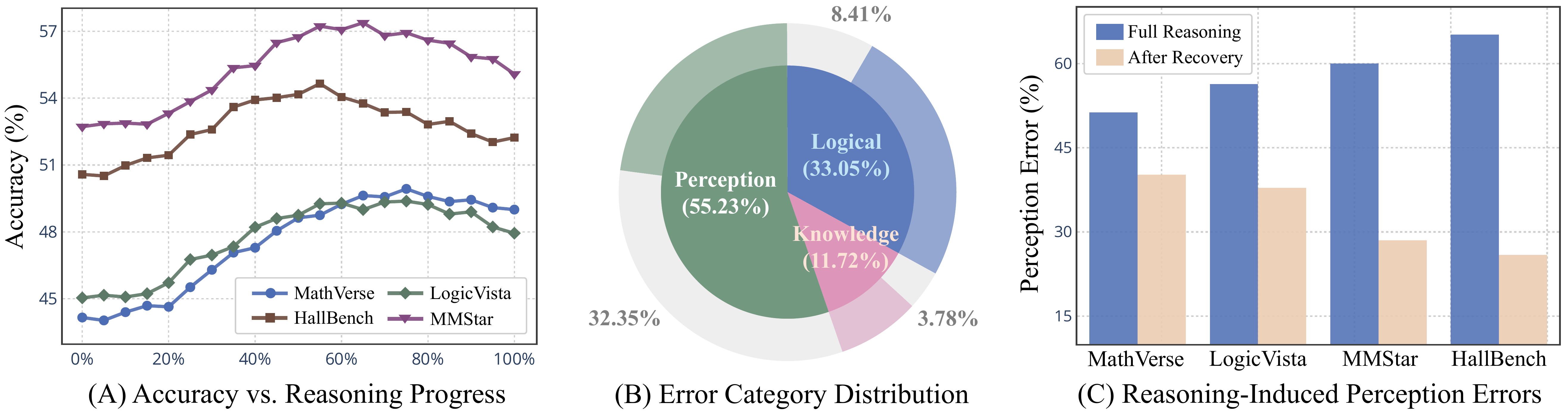}
    \vspace{-6mm}
    \caption{The dual nature analysis of reasoning. In (A), we depict how accuracy evolves throughout the reasoning process. In (B), we show the distribution of error categories under full reasoning (inner ring), alongside the proportion recoverable via early decision (outer ring). In (C), we further present the initial ratio of perception errors on different benchmarks (left bar), as well as the remaining errors that persist after recovery (right bar). It is important to note that recoverable ratio is a loose metric based on a series of early decisions along the reasoning trajectory rather than a single evaluation.
    }
\label{fig:perception}
\vspace{-3mm}
\end{figure*}

We first explore how accuracy evolves throughout the reasoning process in existing multimodal reasoning models. Specifically, we select three representitve VLMs, \ie, R1-OneVision~\citep{yang2025r1}, VLAA-Thinker~\citep{chen2025sft} and Vision-R1~\citep{huang2025vision}, spanning both SFT-then-RL and RL-only training pipelines. We evaluate models on reasoning-centric benchmarks encompassing MathVerse~\citep{zhang2024mathverse} and LogicVista~\citep{xiao2024logicvista}, and vision-intensive benchmarks including MMStar~\citep{chen2024we} and HallusionBench~\citep{guan2024hallusionbench}. To enable fine-grained analysis, we introduce an early decision mode, in which the model is prompted to terminate reasoning at intermediate logical boundaries such as commas, periods or line breaks, and provide an immediate answer, as illustrated in Fig.~\ref{fig:motivation}. This setting allows us to quantify the contribution of each reasoning segment to the overall task performance.

\textbf{Longer reasoning does not guarantee better performance}. As shown in Fig.~\ref{fig:perception}~(A), we leverage early decision to control the reasoning length and monitor accuracy trend while reasoning progresses from $0\%$, \ie, direct answering to $100\%$, \ie, full reasoning. In the early stages, reasoning yields substantial gains across benchmarks, consistent with prior works. However, as reasoning continues, these gains tend to plateau and even begin to reverse, leading to a decline in accuracy. This dual effect is particularly pronounced on benchmarks such as MMStar and HallusionBench, where accuracy drops by over $2\%$ from the peak, almost offsetting the original benefits of reasoning.

The above analysis suggests that, while reasoning is overall effective, its potential is constrained by a hidden factor that acts as a significant bottleneck. To further examine this limitation, we perform a comprehensive error analysis. In particular, we classify incorrect responses of Vision-R1 under full reasoning into three categories, based on the characteristics of their reasoning trajectories:

1) \textbf{Perception Error}. Reasoning demonstrates notable deficiencies in the understanding and recognition of visual details such as chart misreading or hallucination exemplified in Fig.~\ref{fig:motivation}.

2) \textbf{Logical Error}. Models exhibit symbolic reasoning errors such as arithmetic inaccuracies, logically invalid inferences, or incoherence across intermediate reasoning steps.

3) \textbf{Knowledge Error}. There are commonsense violations or factual errors that may stem from outdated knowledge and inherent limitations in the pre-training phase.

We employ GPT-5~\citep{Singh_2025} to identify error categories and quantify their distribution, with details provided in Appendix~\hyperref[subsec:A3]{A.3}. Beyond this, we further investigate the impact of reasoning progress on these failure cases. Specifically, we apply early decision, where an error case is considered recoverable if the model can produce a correct answer at any earlier point along the reasoning trajectory, and report the corresponding proportion of such cases. These recoverable instances suggest that the model is initially on the right path but is ultimately misled by prolonged reasoning.

\textbf{The harder the model thinks, the worse the model sees}. As shown in Fig.~\ref{fig:perception}~(B), despite the complexity of these reasoning benchmarks, the majority of errors made by VLMs stem not from logical failures ($33.05\%$), but rather from basic perception errors ($55.23\%$). Remarkably, a substantial portion of these perceptual errors ($32.35\%$) may be recovered by terminating the reasoning process in advance through early decision. This counterintuitive result suggests that as reasoning progresses, the model's perceptual capability gradually weakens. Together with the finding in Fig.~\ref{fig:perception}~(A), these degradation in visual perception caused by prolonged thoughts is most likely a key contributor to the performance decline observed in the later stages of reasoning.

\textbf{The harms of reasoning are most evident in vision-heavy tasks}. In Fig.~\ref{fig:perception}~(C), we further examine the impact of reasoning progression on perception errors across different benchmarks. Notably, in MMStar and HallusionBench, which feature high-resolution real-world images and perceptually elusive content, the proportion of perception errors arising during reasoning is substantially higher. Moreover, the share of recoverable instances via early decision in these benchmarks is markedly greater than that observed in others, such as MathVerse and LogicVista. This observation suggests that, in contrast to tasks with simple visual structures, \eg, geometry, the shortcomings of reasoning become more pronounced when applied to vision-intensive problems. 

\vspace{-1mm}
\subsection*{3.2 \quad Side Effect of Reasoning: Visual Forgetting}
\label{subsec:3.2}

\begin{figure*}[t!] 
    \centering
    \includegraphics[width=1\linewidth]{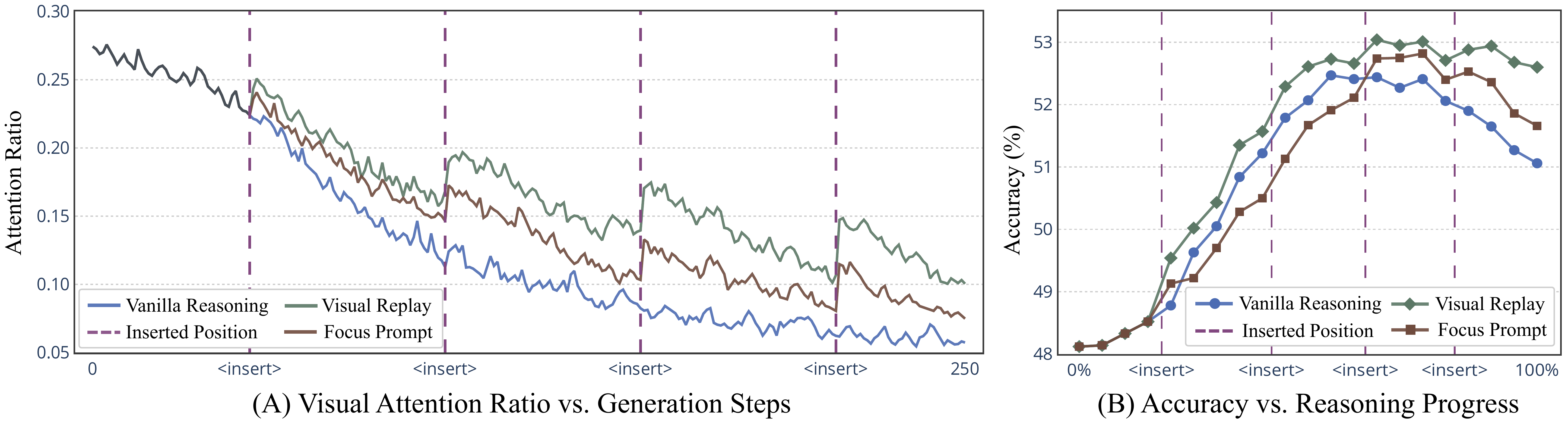}
    \vspace{-6mm}
    \caption{The analysis of visual forgetting. In (A), we randomly select an example from MathVerse and visualize the evolution of attention ratio to visual tokens from Vision-R1 across generation steps. We consider vanilla reasoning as well as two variants, \ie, visual replay and focus prompt. In (B), we report the average accuracy across established benchmarks and models, and track its trend throughout the reasoning process under the aforementioned three settings. The average cutoff positions across examples to insert images or instructions are indicated in vertical dash lines.
    }
\label{fig:forget}
\vspace{-6mm}
\end{figure*}

The above findings highlight a severe trade-off between logical capability and perceptual grounding in multimodal reasoning, which has emerged as a major bottleneck in the development of VLMs. Therefore, to further probe the underlying causes, we track the evolution of attention ratio assigned to visual tokens at each generation step, which provides a proxy of how visual information contributes to the reasoning process. As shown in Fig.~\ref{fig:forget}~(A), under vanilla reasoning, the attention of visual tokens exhibits a marked decline as reasoning progresses, eventually reaching negligible levels. This pattern indicates the model's decision-making becomes increasingly driven by its own historical thoughts rather than the visual cues, a phenomenon termed visual forgetting. To validate this hypothesis, we design two straightforward inference-level remedies:

1) \textbf{Visual Replay}. Instead of presenting the visual input only at the beginning, we reintroduce the image to models at regular intervals throughout the reasoning process.

2) \textbf{Focus Prompt}. Similarly, at regular intervals, we explicitly prompt models to revisit the input image with instructions such as \enquote{$\rm I \ need \ to \ see \ the \ image$} or \enquote{$\rm I \ have \ to \ look \ back$}.

\textbf{Encouraging models to look more often boosts reasoning}. Back to Fig.~\ref{fig:forget}~(A), we insert either the image or instruction at four positions along the reasoning trajectory, which are approximately evenly spaced and aligned with logical boundaries; see Appendix~\hyperref[subsec:A5]{A.5} for more details. Notably, both approaches trigger a sharp and immediate increase in visual attention upon insertion. As shown in Fig.~\ref{fig:forget}~(B), this reinforcement of visual information alleviates the performance degradation observed during the reasoning process, with visual replay yielding particularly pronounced gains, ultimately outperforming vanilla reasoning by around $1.5\%$. This also empirically shows that visual forgetting is the fundamental cause preventing reasoning from realizing its full potential in multimodal tasks.

\section*{4 \quad The Proposed Method}

\vspace{-1mm}
\subsection*{4.1 \quad Preliminary}
\label{subsec:4.1}

To introduce our method, we first revisit the standard RL-based algorithm, typically Group Relative Policy Optimization (GRPO), which is a simplified variant of Proximal Policy Optimization (PPO) that eliminates the critic model. In the multimodal context, consider a dataset $\mathcal{D}$, where each instance consists of a question $q$, a ground truth answer $y$, and a visual input $\mathcal{I}$. The objective of GRPO is to encourage high-quality responses while penalizing inferior ones by comparing a set of generated candidates. Formally, given a policy model $\pi_{\theta}$, we aim to maximize the following:

{\setlength{\abovedisplayskip}{-10pt}
\setlength{\belowdisplayskip}{4pt}
\begin{alignat}{2}
&\mathcal{J}_{\rm GRPO}(\theta) = 
\mathbb{E}_{(q, y, \mathcal{I})\sim \mathcal{D}, \{o_i\}^{G}_{i=1}\sim {\pi}_{\theta_{\rm old}}(\cdot \mid q, \mathcal{I})} \nonumber \\
& \qquad \left[\frac{1}{G}\sum_{i=1}^{G}\frac{1}{\left| o_{i} \right|}\sum_{t=1}^{\left| o_{i} \right|} 
\biggl(\min\Bigl(r_{i, t}(\theta)\hat{A}_{i, t},
{\rm clip}\Bigl(r_{i, t}(\theta), 1-\epsilon, 1+\epsilon\Bigr)\hat{A}_{i, t}\Bigr)
- \lambda D_{\rm KL}\left(\pi_\theta \mid \mid \pi_{\rm ref}\right)
\biggr)\right] \nonumber
\\ 
&{\rm with} \ r_{i, t}(\theta) = \frac{\pi_{\theta}(o_{i,t}\mid q, \mathcal{I}, o_{i, <t})}{\pi_{\theta_{\rm old}}(o_{i,t}\mid q, \mathcal{I}, o_{i, <t})}.
\end{alignat}}

$G$ denotes the number of sequences $o_i$ sampled from the old policy $\pi_{\rm old}$. $\epsilon$ and $\lambda$ are hyperparameters that constrain the clipping bounds to ensure on-policy training and penalize deviations from the reference policy $\pi_{\rm ref}$ to stabilize optimization. The estimated token-level advantage $\hat{A}_{i, t}$ is derived by broadcasting the normalized sequence-level reward $R_{i}$, which is defined as follows:

{\setlength{\abovedisplayskip}{-10pt}
\setlength{\belowdisplayskip}{0pt}
\begin{align}
    \hat{A}_{i,t} = \frac{R_i - {\rm mean}\left(\mathbf{R}\right)}{{\rm std}\left(\mathbf{R}\right)}, \quad i=1,\cdots,G,
\end{align}}

where $\mathbf{R}=\{R_1, R_2, ..., R_{G}\}$ indicates the reward of the sequence group. The design of reward functions offers considerable flexibility. A widely adopted approach involves the use of verifiable rewards, which evaluate model responses by comparing them to ground-truth answers to derive accuracy-based and format-consistency signals.

\vspace{-1mm}
\subsection*{4.2 \quad Vision-Anchored Policy Optimization}
\begin{figure*}[t!] 
    \centering
    \includegraphics[width=1\linewidth]{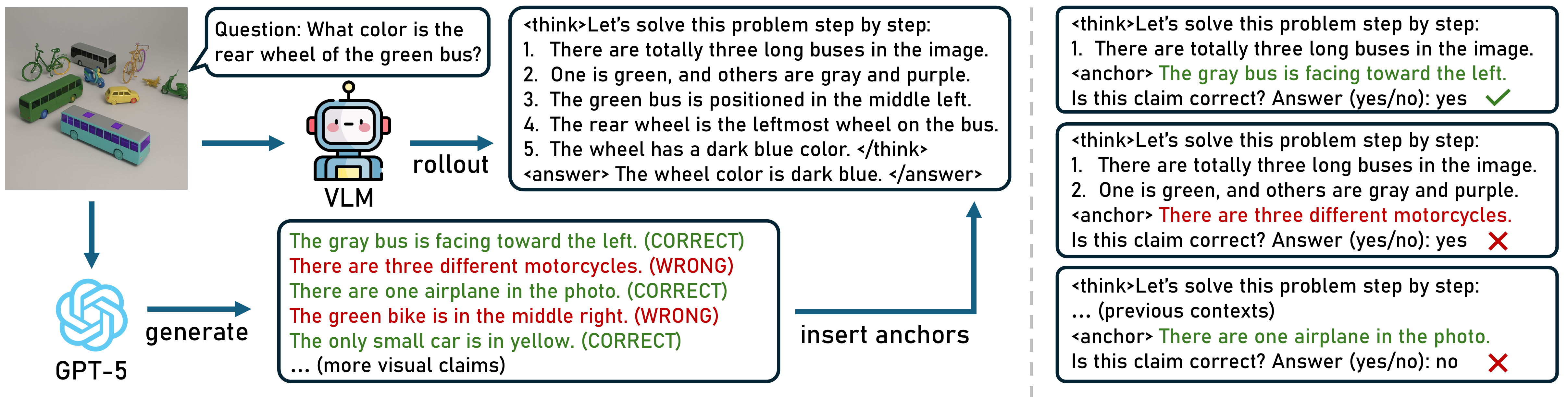}
    \vspace{-7mm}
    \caption{The overview of \texttt{VAPO}. On the left, we first employ GPT to generate a set of claims about visual input, each of which may be either factually correct or not. These claims are then treated as anchors and inserted into the model's reasoning process. Specifically, on the right, for each anchor, we randomly sample a prefix from the reasoning trajectory, append the claim to this truncated context, and probe the model judgment regarding the claim's validity. 
    }
\label{fig:method}
\vspace{-4mm}
\end{figure*}

Although GRPO has become a common practice for training reasoning models, our findings in Section~\hyperref[sec:3]{3} reveal that existing VLMs suffer from visual forgetting, constituting a major bottleneck in multimodal reasoning. To address this, we propose {\sc Vision-Anchored Policy Optimization} (\texttt{VAPO}), a simple yet effective approach as a multimodal replacement for GRPO that steers the reasoning process toward visually grounded trajectories. As shown in Fig.~\ref{fig:method}, \texttt{VAPO} first generates a set of correct and incorrect claims that describe the visual details in the images. These claims are strategically inserted as visual anchors along the reasoning path, where the model is prompted to judge their veracity, thereby enabling ongoing evaluation of model's perceptual capability throughout the reasoning process. Consequently, we introduce a perception reward, aggregated over all anchor points, which explicitly incentivizes the model to retain and utilize visual cues during reasoning.

\textbf{Generating Visual Claims}. We use visual claims as a proxy to evaluate the model's perceptual capability at various reasoning stages. To serve this purpose, the claims need to satisfy two criteria: 1) balanced, \ie, the set should contain an equal number of correct and wrong claims to avoid biased evaluation and 2) independent, \ie, the model judgment of each claim must rely solely on visual input, rather than historical reasoning. Specifically, for each example, we employ GPT-5 to generate a diverse set of claims with balanced correctness. To enforce independence from reasoning contexts, we only provide visual input without revealing the corresponding question, thereby ensuring claim verification is grounded purely in perceptual understanding. See Appendix~\hyperref[subsec:A6]{A.6} for more details.

\textbf{Inserting Visual Anchors}. Upon obtaining these claims, we set up a series of visual anchors within the model's reasoning process, functioning as intermediate checkpoints. Once reaching the anchor, the model is queried with a selected claim to assess its perceptual capability at that stage of reasoning. Formally, given a trajectory $o_i=(o_{i, 1}, o_{i, 2}, \cdots,o_{i, T})$, we define a set of anchors $\mathcal{A}_{i}=\{a_1,a_2,\cdots,a_K\}$ with $a_k \in [1, T]$ denoting the position randomly distributed throughout the trajectory. At each anchor $a_k$, a claim $c_k \in \mathcal{C}_{i}$ is sampled from the corresponding claim pool and appended to the prefix reasoning context, yielding a binary score over the judgement:

{\setlength{\abovedisplayskip}{-10pt}
\setlength{\belowdisplayskip}{2pt}
\begin{align}
    s_k=\mathbf{1}\left[\arg \max_{j\in\{\rm yes, no\}}\pi_{\theta}(j\mid q,\mathcal{I},o_{i, <a_k},c_{k})=l_{k}\right],
\end{align}}

where $l_k$ denotes the ground truth of the claim. In other words, we evaluate the model's perceptual capability by probing its binary decision on the claim and scoring it against the reference label.

\textbf{Perception Reward}. Building on the anchors above, we introduce a perception reward that quantifies the model's overall perceptual capability throughout the reasoning process. Formally, given the per-anchor scores $\{s_k\}_{k=1}^{K}$, we design a late-emphasis weighted aggregation:

{\setlength{\abovedisplayskip}{-8pt}
\setlength{\belowdisplayskip}{2pt}
\begin{align}
    R_{\rm perc} = \frac{\sum_{k=1}^{K}w_{k}s_{k}}{\sum_{k=1}^{K}w_{k}} \quad {\rm with} \ \ w_{k} = \exp\left(\beta\cdot\frac{a_k}{T}\right),
\end{align}}

where $w_k$ indicates the weight assigned to the corresponding anchor, and $\beta$ is a hyperparameter that controls the degree of emphasis of later anchors. This design is motivated by the observation that the model's perceptual capability tends to decline as reasoning progresses. Consequently, greater weight is assigned to anchors in the latter stages to precisely target the model weakness. In summary, the final reward for a sampled sequence $o_{i}$ is computed as:
{\setlength{\abovedisplayskip}{7pt}
\setlength{\belowdisplayskip}{2pt}
\begin{align}
    R_{i} = R_{\rm acc} + R_{\rm fmt} + \gamma \cdot \mathbf{1}\left[R_{\rm acc} = 1\right]\cdot R_{\rm perc},
\end{align}}

where $R_{\rm acc}$ and $R_{\rm fmt}$ denote accuracy and format reward, $\gamma$ is the weight of perception reward. In practice, we impose an accuracy condition on perception reward to guard against reward hacking, thereby preventing models from simply boosting perceptual capability by producing trivially short reasoning paths. Our training procedure follows the objective of GRPO as defined in Section~\hyperref[subsec:4.1]{4.1}.

\vspace{-1mm}
\section*{5 \quad Experiment and Results}
\textbf{Implementation Details}. Unless otherwise specified, all models are trained on ViRL39K~\citep{vl-rethinker}, a high-quality and comprehensive dataset, for $2$ epochs with a learning rate of $5\rm e^{-6}$. We adopt Qwen2.5-VL~\citep{bai2025qwen2} with 3B and 7B parameters as base models. For group reward computation, we follow previous works~\citep{huang2025vision, chen2025sft} by generating $5$ responses per example with a sampling temperature of $1.0$. For \texttt{VAPO}, we set the default number of anchors to $K=20$, the late-emphasis weight to $\beta=1.5$, and reward weight $\gamma = 0.1$. 

\textbf{Benchmarks}. We denote the models trained with our method as \texttt{VAPO-Thinker}, and evaluate on ten benchmarks covering diverse task types: mathmatical benchmarks such as MathVerse~\citep{zhang2024mathverse}, MathVista~\citep{lu2023mathvista}, MathVision~\citep{wang2024measuring}, LogicVista~\citep{xiao2024logicvista}, WeMath~\citep{qiao2024we} and Geometry3k~\citep{lu2021inter}, as well as general-purpose ones including MMMU~\citep{yue2024mmmu}, MMStar~\citep{chen2024we}, HallusionBench~\citep{guan2024hallusionbench} and MMVet~\citep{yu2023mm}. Further evaluation results are provided in Appendix~\hyperref[sec:B]{B}.

\textbf{Baseline Methods}. To verify the effectiveness of our approach, we compare \texttt{VAPO-Thinker} with a range of existing strong reasoning models, including proprietary ones such as GPT-5-Thinking~\citep{Singh_2025} and Gemini-2.5-Pro~\citep{comanici2025gemini}, as well as open-source counterparts encompassing InternVL2.5~\citep{chen2024expanding}, R1-OneVision~\citep{yang2025r1}, VLAA-Thinker~\citep{chen2025sft}, and Vision-R1~\citep{huang2025vision}. The baseline results are primarily referenced from the corresponding papers, secondarily from the OpenCompass leaderboard, and reproduced when neither source is available. See Appendix~\hyperref[subsec:A7]{A.7} for more configuration details.

\vspace{-1mm}
\subsection*{5.1 \quad Main Results}
\begin{table*}[t]
\footnotesize
\centering
\setlength{\tabcolsep}{1.6mm}
\setlength\heavyrulewidth{0.2ex}
\renewcommand{\arraystretch}{1.0}
\begin{NiceTabular}{@{}lccccccc@{}}
\toprule
\multicolumn{1}{c}{Models}   & MathVerse            & MathVista            & MathVision           & LogicVista           & WeMath               & Geometry3k                & Avg.              \\ \midrule
\textit{Close-source models} & \multicolumn{1}{l}{} & \multicolumn{1}{l}{} & \multicolumn{1}{l}{} & \multicolumn{1}{l}{} & \multicolumn{1}{l}{} & \multicolumn{1}{l}{} & \multicolumn{1}{l}{} \\
GPT-5-Thinking*               & 81.2                 & 81.9                 & 72.0                 & 70.0                 & 71.1                 & 79.9                 & 76.1                 \\
Gemini-2.5-Pro*               & 76.9                 & 80.9                 & 69.1                 & 73.8                 & 78.0                 & 77.2                 & 75.9                 \\ \midrule
\textit{Open-source models}  & \multicolumn{1}{l}{} & \multicolumn{1}{l}{} & \multicolumn{1}{l}{} & \multicolumn{1}{l}{} & \multicolumn{1}{l}{} & \multicolumn{1}{l}{} & \multicolumn{1}{l}{} \\
Qwen2.5-VL-7B                & 40.7                 & 62.3                 & 23.2                 & 42.6                 & 33.1                 & 38.5                 & 40.1                 \\
InternVL2.5-8B*               & 34.5                 & 68.2                 & 25.6                 & 38.3                 & 38.6                 & 44.8                 & 41.7                 \\
R1-OneVision-7B             & 46.4                 & 64.1                 & 29.9                 & 45.6                & \textbf{44.6}        & 46.1                 & 46.1                 \\
VLAA-Thinker-7B             & 48.2                 & 68.0                 & 26.4                 & 48.5                 & 41.5                 & 50.6                 & 47.2                 \\
Vision-R1-7B                & 52.4                 & 73.5                 & 28.2                 & 49.7                 & 41.6                 & 49.0                 & 49.1                 \\ \midrule
\textit{Our models}          & \multicolumn{1}{l}{} & \multicolumn{1}{l}{} & \multicolumn{1}{l}{} & \multicolumn{1}{l}{} & \multicolumn{1}{l}{} & \multicolumn{1}{l}{} & \multicolumn{1}{l}{} \\
\texttt{VAPO-Thinker-3B}              & 35.8                 & 67.1                 & 23.9                 & 39.7                 & 35.4                 & 44.2                & 41.0                 \\
\rowcolor[HTML]{EDEDED} \texttt{VAPO-Thinker-7B}              & \textbf{53.3}                 & \textbf{75.6}        & \textbf{31.9}        & \textbf{50.9}        & 43.6                 & \textbf{51.3}        & \textbf{51.1}        \\ \bottomrule
\end{NiceTabular}
\vspace{-3mm}
\caption{The evaluation of our proposed method on various mathematical benchmarks. We report results of existing open-source multi-modal reasoning models, as well as proprietary models for reference. Note that * indicates baseline results referenced from the OpenCompass leaderboard.}
\vspace{-1mm}
\label{tab:math}
\end{table*}

\begin{figure}[!t]
\begin{minipage}[t]{\textwidth}
\begin{minipage}[t]{0.49\textwidth}
\makeatletter\def\@captype{table}
\footnotesize
\setlength{\tabcolsep}{1.28mm}
\setlength\heavyrulewidth{0.2ex}
\renewcommand{\arraystretch}{1.0}
\begin{NiceTabular}{@{}lccccc@{}}
\toprule
\multicolumn{1}{c}{Models} & MMMU          & MMStar        & Hall     & MMVet         & Avg.       \\ \midrule
Qwen-VL                & 52.7          & 54.9          & 50.0          & 64.8          & 55.6          \\
R1-OV               & 54.3          & 54.1          & 52.5          & 65.2          & 56.5          \\
VLAA               & 59.1          & 49.7          & 54.7          & 70.0          & 58.4          \\
V-R1                  & 57.6          & 61.4          & 49.5          & 71.1          & 59.9          \\ 
\rowcolor[HTML]{EDEDED} \texttt{VAPO}               & \textbf{60.2} & \textbf{63.0} & \textbf{57.4} & \textbf{71.9} & \textbf{63.1} \\ \bottomrule
\end{NiceTabular}
\vspace{-3mm}
\caption{The evaluation of our proposed method on general-purpose benchmarks. All baselines considered in this evaluation are of the 7B scale.}
\label{tab:general}
\end{minipage}
\hspace*{0.15cm}
\begin{minipage}[t]{0.49\textwidth}
\makeatletter\def\@captype{table}
\footnotesize
\setlength{\tabcolsep}{1.45mm}
\setlength\heavyrulewidth{0.2ex}
\renewcommand{\arraystretch}{1.0}
\begin{NiceTabular}{@{}lccccc@{}}
\toprule
\multicolumn{1}{c}{Models} & WeMath          & Geo3k       & \multicolumn{1}{l}{MMStar} & \multicolumn{1}{l}{Hall} & Avg.          \\ \midrule
VLAA\textsubscript{FP}                  & 41.8          & 50.7          & 51.1                      & 55.2                       & 49.7          \\
VLAA\textsubscript{VR}                  & 42.3          & 51.1          & 52.9                      & 56.2                       & 50.6          \\
V-R1\textsubscript{FP}                  & 42.1          & 49.7          & 61.8                      & 50.5                       & 51.0          \\
V-R1\textsubscript{VR}                  & 42.5          & 50.5          & 62.1                      & 51.8                       & 51.7          \\ 
\rowcolor[HTML]{EDEDED} \texttt{VAPO}                     & \textbf{43.6} & \textbf{51.3} & \textbf{63.0}             & \textbf{57.4}              & \textbf{53.8} \\ \bottomrule
\end{NiceTabular}
\vspace{-3mm}
\caption{Comparison with strong baselines augmented with test-time remedies. FP and VR denote focus prompt and visual replay, respectively.}
\label{tab:inference}
\end{minipage}
\vspace{-0.7em}
\end{minipage}
\vspace{-2mm}
\end{figure}

\textbf{\texttt{VAPO} consistently improves accuracy across diverse benchmarks}. As shown in Table~\ref{tab:math}, \texttt{VAPO-Thinker-7B} outperforms recent reasoning models of the same scale on mathematical problems, achieving an average improvement of $2\%$ ($49.1\% \rightarrow 51.1\%)$. The advantage is more pronounced on general-purpose tasks, as shown in Table~\ref{tab:general}, where our method surpasses previous best results by $3.2\%$ ($59.9\% \rightarrow 63.1\%$), thereby establishing a new state of the art. This suggests that compared to logic-heavy problems such as math, \texttt{VAPO} is particularly effective on vision-intensive tasks like MMStar and HallusionBench, demonstrating stronger visually grounded reasoning.

\textbf{\texttt{VAPO} offers a principled solution beyond inference-level remedies}. In Section~\hyperref[subsec:3.2]{3.2}, we introduce two remedies, \ie, visual replay and focus prompt, as preliminary exploration to verify the visual forgetting issue. Here we compare \texttt{VAPO} against these test-time strategies. As shown in Table~\ref{tab:inference}, even when strong baselines are equipped with visual replay or focus prompt, \texttt{VAPO} still achieves substantially higher performance, suggesting that inference-level remedies alone are insufficient to address visual forgetting, whereas \texttt{VAPO} fundamentally rectifies this reasoning deficiency.

\vspace{-1mm}
\subsection*{5.2 \quad Further Discussion}

\begin{figure*}[t!] 
    \centering
    \includegraphics[width=1\linewidth]{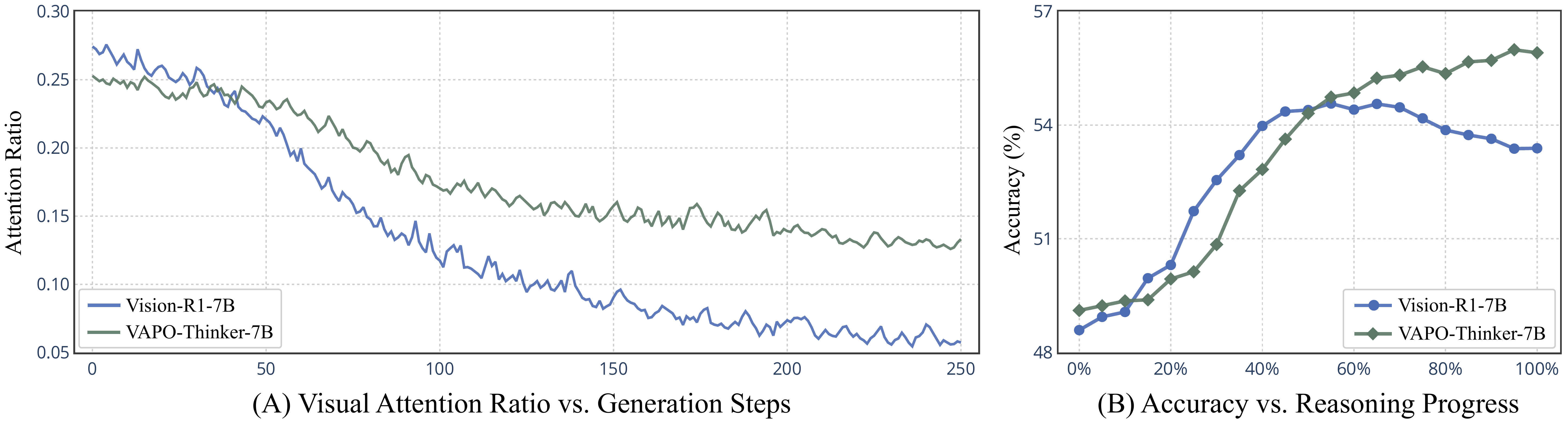}
    \vspace{-7mm}
    \caption{Zoom-in analysis of our method. In (A), we compare the evolution of visual attention across generation steps between \texttt{VAPO} and the baseline on the selected example. In (B), we track the average accuracy across ten benchmarks throughout the reasoning process via early decision.
    }
    \vspace{-2mm}
    \label{fig:result}
\end{figure*}

\begin{figure}[!t]
\begin{minipage}[t]{\textwidth}
\begin{minipage}[t]{0.497\textwidth}
\makeatletter\def\@captype{figure}
\includegraphics[width=1\linewidth]{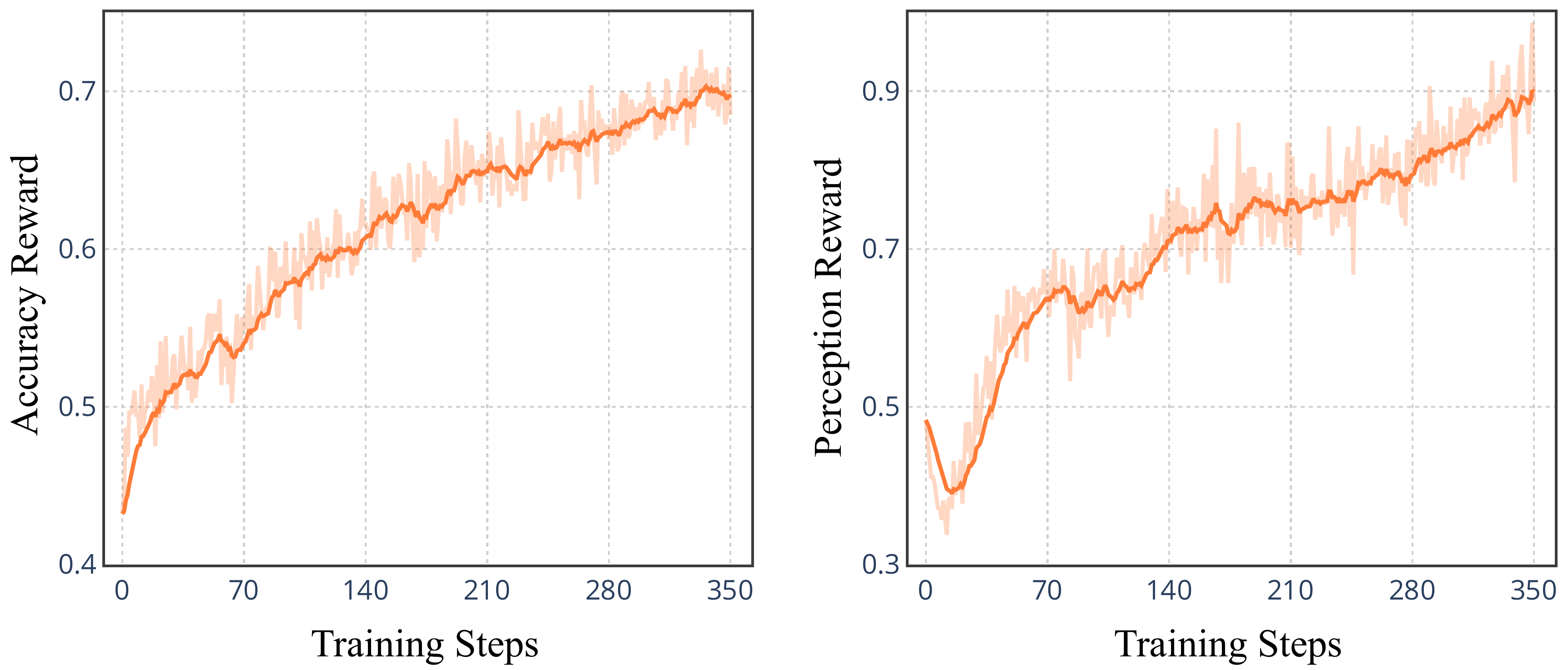}
\vspace{-7mm}
\caption{The learning curves of accuracy and perception reward during the training stage.}
\label{fig:dynamic}
\end{minipage}
\hspace*{0.15cm}
\begin{minipage}[t]{0.48\textwidth}
\makeatletter\def\@captype{figure}
\includegraphics[width=1\linewidth]{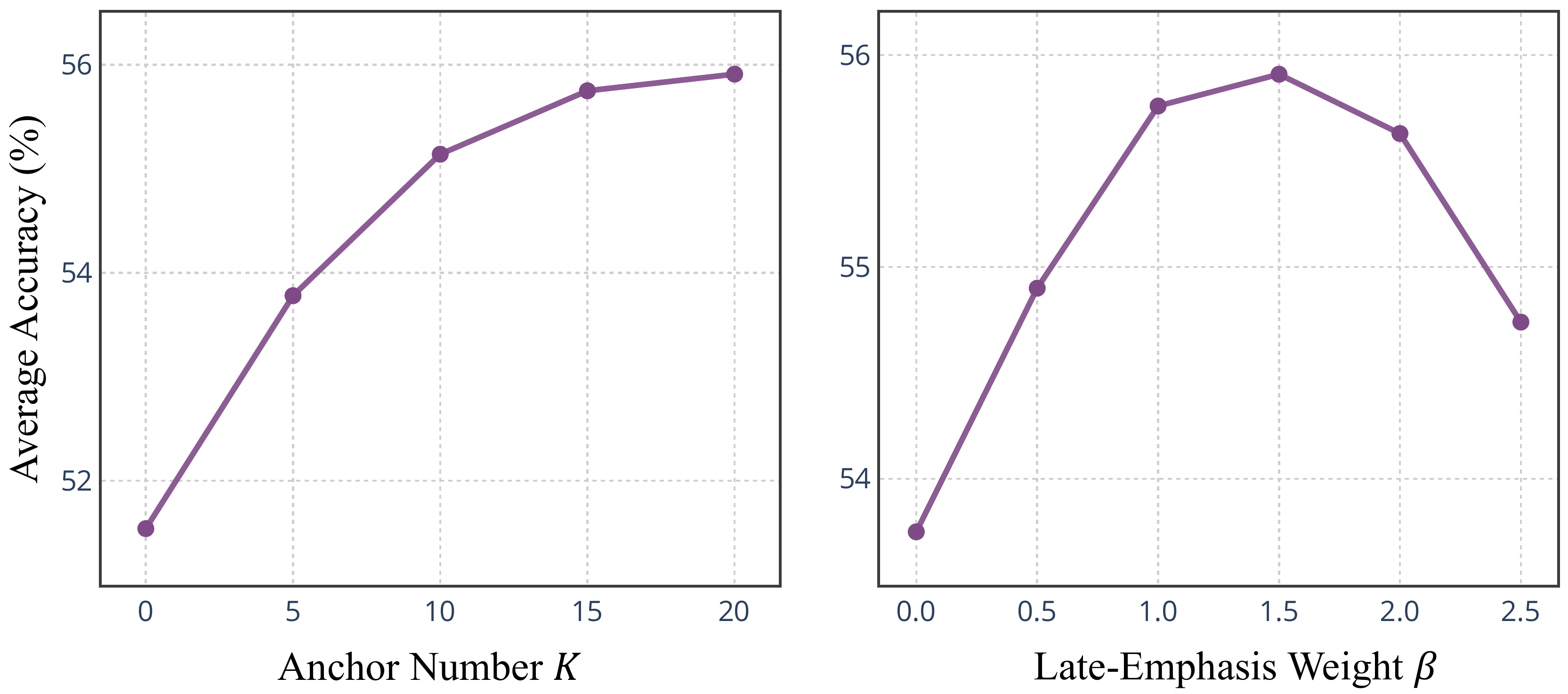}
\vspace{-7mm}
\caption{The ablation study on the effects of anchor number $K$ and late-emphasis weight $\beta$.}
\label{fig:ablation}
\end{minipage}
\vspace{-0.7em}
\end{minipage}
\vspace{-2mm}
\end{figure}

\textbf{\texttt{VAPO} fully releases the potentials of reasoning}. Here we analyze the evolution of visual attention and accuracy over the course of reasoning process. In Fig.~\ref{fig:result}~(A), compared with the baseline, our model demonstrates a more gentle decline in the attention ratio, sustaining a consistently higher level in the later stages, indicating that \texttt{VAPO} effectively strengthens the contribution of visual cues to model decisions. The benefit brought by this is directly reflected in accuracy, as shown in Fig.~\ref{fig:result}~(B), where in contrast to the baseline which exhibits a sharp performance decline during later steps, our method achieves steadily increasing accuracy, thereby fully leveraging the advantages of reasoning.

\textbf{\texttt{VAPO} greatly enhances perceptual capability during reasoning}. In Fig.~\ref{fig:dynamic}, we plot the training curves of accuracy and perception reward. The results reveal that perception reward exhibits a steady upward trend in parallel with accuracy, suggesting that the model progressively allocates greater attention to visual input and thereby improves its ability to verify visual claims. Interestingly, an sharp decline in perception reward is observed during the early stage, which may be attributed to the initial dominance of accuracy and format alignment, causing the model to temporarily disregard visual information. This underscores a fundamental limitation of prior standard training paradigms.

\vspace{-1mm}
\subsection*{5.3 \quad Ablation Study}
\textbf{More anchors yield better visually grounded reasoning}. We examine the impact of visual anchor number $K$ on average accuracy across ten benchmarks. As shown on the left of Fig.~\ref{fig:ablation}, we vary $K$ from $0$, which degrades to vanilla GRPO training, to $20$, which is adopted as our default configuration. The results indicate that the average accuracy improves rapidly as $K$ increases and approaches saturation at around $K=20$. This behavior is intuitive, as setting up denser anchors during reasoning provides a more reliable assessment of the model's perceptual capability.

\textbf{Later anchors contribute to greater gains}. In \texttt{VAPO}, the parameter $\beta$ controls the relative emphasis placed on later anchors to target longer reasoning. As shown on the right of Fig.~\ref{fig:ablation}, we evaluate the effect of varying $\beta$ on average accuracy. When $\beta=0$, all anchors are assigned equal weights, while larger values assign greater importance to later anchors. The results show that performance peaks at $\beta=1.5$, where around $50\%$ of the total weight is concentrated on the last $30\%$ of anchors. More results such as ablation of reward weight $\gamma$ and limitation analysis are provided in Appendix~\hyperref[sec:B]{B}.


\vspace{-1mm}
\section*{6 \quad Conclusion}
In this study, we conduct an investigation into the capabilities and limitations of multimodal reasoning. We first reveal the dual nature of reasoning: while it enhances logical inference and proves beneficial for challenging problems, longer reasoning length may gradually impair perceptual grounding, leading to recognition deficiencies on otherwise basic visual questions. Then we further identify this phenomenon and attribute it to visual forgetting, where prolonged reasoning length causes the model to disregard visual information. To address this, we propose {\sc Vision-Anchored Policy Optimization} (\texttt{VAPO}), a simple yet effective method that steers the reasoning process along trajectories anchored in visual evidence. Our result model, \texttt{VAPO-Thinker-7B}, effectively mitigates visual forgetting and establishes a new state of the art across various benchmarks.

\subsubsection*{Acknowledgement}
This research was, in part, funded by the U.S. Government – DARPA TIAMAT HR00112490421. The views and conclusions contained in this document are those of the authors and should not be interpreted as representing the official policies, either expressed or implied, of the U.S. Government. We also gratefully acknowledge Lambda GPU Cloud and Maincode for their generous provision of computational resources.

\bibliography{iclr2026_conference}

@String(CVPR= {IEEE Conf. Comput. Vis. Pattern Recog.})

@String(ECCV= {Eur. Conf. Comput. Vis.})

@String(ICLR = {Int. Conf. Learn. Represent.})

@String(CVPR  = {CVPR})

@String(ECCV  = {ECCV})

@String(ICLR  = {ICLR})

@article{wei2022chain,
  title={Chain-of-thought prompting elicits reasoning in large language models},
  author={Wei, Jason and Wang, Xuezhi and Schuurmans, Dale and Bosma, Maarten and Xia, Fei and Chi, Ed and Le, Quoc V and Zhou, Denny and others},
  journal={NeurIPS},
  volume={35},
  pages={24824--24837},
  year={2022}
}

@article{kojima2022large,
  title={Large language models are zero-shot reasoners},
  author={Kojima, Takeshi and Gu, Shixiang Shane and Reid, Machel and Matsuo, Yutaka and Iwasawa, Yusuke},
  journal={NeurIPS},
  volume={35},
  pages={22199--22213},
  year={2022}
}

@article{guo2025deepseek,
  title={Deepseek-r1: Incentivizing reasoning capability in llms via reinforcement learning},
  author={Guo, Daya and Yang, Dejian and Zhang, Haowei and Song, Junxiao and Zhang, Ruoyu and Xu, Runxin and Zhu, Qihao and Ma, Shirong and Wang, Peiyi and Bi, Xiao and others},
  journal={arXiv preprint arXiv:2501.12948},
  year={2025}
}

@article{muennighoff2025s1,
  title={s1: Simple test-time scaling},
  author={Muennighoff, Niklas and Yang, Zitong and Shi, Weijia and Li, Xiang Lisa and Fei-Fei, Li and Hajishirzi, Hannaneh and Zettlemoyer, Luke and Liang, Percy and Cand{\`e}s, Emmanuel and Hashimoto, Tatsunori},
  journal={arXiv preprint arXiv:2501.19393},
  year={2025}
}

@article{yang2025r1,
  title={R1-onevision: Advancing generalized multimodal reasoning through cross-modal formalization},
  author={Yang, Yi and He, Xiaoxuan and Pan, Hongkun and Jiang, Xiyan and Deng, Yan and Yang, Xingtao and Lu, Haoyu and Yin, Dacheng and Rao, Fengyun and Zhu, Minfeng and others},
  journal={arXiv preprint arXiv:2503.10615},
  year={2025}
}

@article{huang2025vision,
  title={Vision-r1: Incentivizing reasoning capability in multimodal large language models},
  author={Huang, Wenxuan and Jia, Bohan and Zhai, Zijie and Cao, Shaosheng and Ye, Zheyu and Zhao, Fei and Xu, Zhe and Hu, Yao and Lin, Shaohui},
  journal={arXiv preprint arXiv:2503.06749},
  year={2025}
}

@article{shao2024deepseekmath,
  title={Deepseekmath: Pushing the limits of mathematical reasoning in open language models},
  author={Shao, Zhihong and Wang, Peiyi and Zhu, Qihao and Xu, Runxin and Song, Junxiao and Bi, Xiao and Zhang, Haowei and Zhang, Mingchuan and Li, YK and Wu, Yang and others},
  journal={arXiv preprint arXiv:2402.03300},
  year={2024}
}

@article{liu2025visual,
  title={Visual-rft: Visual reinforcement fine-tuning},
  author={Liu, Ziyu and Sun, Zeyi and Zang, Yuhang and Dong, Xiaoyi and Cao, Yuhang and Duan, Haodong and Lin, Dahua and Wang, Jiaqi},
  journal={arXiv preprint arXiv:2503.01785},
  year={2025}
}

@article{shen2025vlm,
  title={Vlm-r1: A stable and generalizable r1-style large vision-language model},
  author={Shen, Haozhan and Liu, Peng and Li, Jingcheng and Fang, Chunxin and Ma, Yibo and Liao, Jiajia and Shen, Qiaoli and Zhang, Zilun and Zhao, Kangjia and Zhang, Qianqian and others},
  journal={arXiv preprint arXiv:2504.07615},
  year={2025}
}

@article{chu2025sft,
  title={Sft memorizes, rl generalizes: A comparative study of foundation model post-training},
  author={Chu, Tianzhe and Zhai, Yuexiang and Yang, Jihan and Tong, Shengbang and Xie, Saining and Schuurmans, Dale and Le, Quoc V and Levine, Sergey and Ma, Yi},
  journal={arXiv preprint arXiv:2501.17161},
  year={2025}
}

@article{li2025think,
  title={Think or not think: A study of explicit thinking in rule-based visual reinforcement fine-tuning},
  author={Li, Ming and Zhong, Jike and Zhao, Shitian and Lai, Yuxiang and Zhang, Haoquan and Zhu, Wang Bill and Zhang, Kaipeng},
  journal={arXiv preprint arXiv:2503.16188},
  year={2025}
}

@article{xia2025visionary,
  title={Visionary-r1: Mitigating shortcuts in visual reasoning with reinforcement learning},
  author={Xia, Jiaer and Zang, Yuhang and Gao, Peng and Li, Yixuan and Zhou, Kaiyang},
  journal={arXiv preprint arXiv:2505.14677},
  year={2025}
}

@inproceedings{wang2020high,
  title={High-frequency component helps explain the generalization of convolutional neural networks},
  author={Wang, Haohan and Wu, Xindi and Huang, Zeyi and Xing, Eric P},
  booktitle={CVPR},
  pages={8684--8694},
  year={2020}
}

@inproceedings{koh2020concept,
  title={Concept bottleneck models},
  author={Koh, Pang Wei and Nguyen, Thao and Tang, Yew Siang and Mussmann, Stephen and Pierson, Emma and Kim, Been and Liang, Percy},
  booktitle={ICML},
  pages={5338--5348},
  year={2020},
  organization={PMLR}
}

@article{yao2023tree,
  title={Tree of thoughts: Deliberate problem solving with large language models},
  author={Yao, Shunyu and Yu, Dian and Zhao, Jeffrey and Shafran, Izhak and Griffiths, Tom and Cao, Yuan and Narasimhan, Karthik},
  journal={NeurIPS},
  volume={36},
  pages={11809--11822},
  year={2023}
}

@article{yang2025qwen3,
  title={Qwen3 technical report},
  author={Yang, An and Li, Anfeng and Yang, Baosong and Zhang, Beichen and Hui, Binyuan and Zheng, Bo and Yu, Bowen and Gao, Chang and Huang, Chengen and Lv, Chenxu and others},
  journal={arXiv preprint arXiv:2505.09388},
  year={2025}
}

@article{cai2024internlm2,
  title={Internlm2 technical report},
  author={Cai, Zheng and Cao, Maosong and Chen, Haojiong and Chen, Kai and Chen, Keyu and Chen, Xin and Chen, Xun and Chen, Zehui and Chen, Zhi and Chu, Pei and others},
  journal={arXiv preprint arXiv:2403.17297},
  year={2024}
}

@article{yu2025dapo,
  title={Dapo: An open-source llm reinforcement learning system at scale},
  author={Yu, Qiying and Zhang, Zheng and Zhu, Ruofei and Yuan, Yufeng and Zuo, Xiaochen and Yue, Yu and Dai, Weinan and Fan, Tiantian and Liu, Gaohong and Liu, Lingjun and others},
  journal={arXiv preprint arXiv:2503.14476},
  year={2025}
}

@article{zheng2025group,
  title={Group sequence policy optimization},
  author={Zheng, Chujie and Liu, Shixuan and Li, Mingze and Chen, Xiong-Hui and Yu, Bowen and Gao, Chang and Dang, Kai and Liu, Yuqiong and Men, Rui and Yang, An and others},
  journal={arXiv preprint arXiv:2507.18071},
  year={2025}
}

@article{snell2024scaling,
  title={Scaling llm test-time compute optimally can be more effective than scaling model parameters},
  author={Snell, Charlie and Lee, Jaehoon and Xu, Kelvin and Kumar, Aviral},
  journal={ICLR},
  year={2025}
}

@article{peng2025skywork,
  title={Skywork r1v: Pioneering multimodal reasoning with chain-of-thought},
  author={Peng, Yi and Wang, Peiyu and Wang, Xiaokun and Wei, Yichen and Pei, Jiangbo and Qiu, Weijie and Jian, Ai and Hao, Yunzhuo and Pan, Jiachun and Xie, Tianyidan and others},
  journal={arXiv preprint arXiv:2504.05599},
  year={2025}
}

@article{tan2025reason,
  title={Reason-rft: Reinforcement fine-tuning for visual reasoning},
  author={Tan, Huajie and Ji, Yuheng and Hao, Xiaoshuai and Lin, Minglan and Wang, Pengwei and Wang, Zhongyuan and Zhang, Shanghang},
  journal={arXiv preprint arXiv:2503.20752},
  year={2025}
}

@article{chen2025sft,
  title={Sft or rl? an early investigation into training r1-like reasoning large vision-language models},
  author={Chen, Hardy and Tu, Haoqin and Wang, Fali and Liu, Hui and Tang, Xianfeng and Du, Xinya and Zhou, Yuyin and Xie, Cihang},
  journal={arXiv preprint arXiv:2504.11468},
  year={2025}
}

@article{meng2025mm,
  title={Mm-eureka: Exploring the frontiers of multimodal reasoning with rule-based reinforcement learning},
  author={Meng, Fanqing and Du, Lingxiao and Liu, Zongkai and Zhou, Zhixiang and Lu, Quanfeng and Fu, Daocheng and Han, Tiancheng and Shi, Botian and Wang, Wenhai and He, Junjun and others},
  journal={arXiv preprint arXiv:2503.07365},
  year={2025}
}

@inproceedings{chen2024image,
  title={An image is worth 1/2 tokens after layer 2: Plug-and-play inference acceleration for large vision-language models},
  author={Chen, Liang and Zhao, Haozhe and Liu, Tianyu and Bai, Shuai and Lin, Junyang and Zhou, Chang and Chang, Baobao},
  booktitle={ECCV},
  pages={19--35},
  year={2024},
  organization={Springer}
}

@inproceedings{fu2024blink,
  title={Blink: Multimodal large language models can see but not perceive},
  author={Fu, Xingyu and Hu, Yushi and Li, Bangzheng and Feng, Yu and Wang, Haoyu and Lin, Xudong and Roth, Dan and Smith, Noah A and Ma, Wei-Chiu and Krishna, Ranjay},
  booktitle={ECCV},
  pages={148--166},
  year={2024},
  organization={Springer}
}

@article{tu2025attention,
  title={Attention reallocation: Towards zero-cost and controllable hallucination mitigation of mllms},
  author={Tu, Chongjun and Ye, Peng and Zhou, Dongzhan and Bai, Lei and Yu, Gang and Chen, Tao and Ouyang, Wanli},
  journal={arXiv preprint arXiv:2503.08342},
  year={2025}
}

@inproceedings{leng2024mitigating,
  title={Mitigating object hallucinations in large vision-language models through visual contrastive decoding},
  author={Leng, Sicong and Zhang, Hang and Chen, Guanzheng and Li, Xin and Lu, Shijian and Miao, Chunyan and Bing, Lidong},
  booktitle={CVPR},
  pages={13872--13882},
  year={2024}
}

@article{wang2024mitigating,
  title={Mitigating hallucinations in large vision-language models with instruction contrastive decoding},
  author={Wang, Xintong and Pan, Jingheng and Ding, Liang and Biemann, Chris},
  journal={ACL Findings},
  year={2024}
}

@article{gong2024damro,
  title={Damro: Dive into the attention mechanism of lvlm to reduce object hallucination},
  author={Gong, Xuan and Ming, Tianshi and Wang, Xinpeng and Wei, Zhihua},
  journal={arXiv preprint arXiv:2410.04514},
  year={2024}
}

@article{wang2025perception,
  title={Perception-aware policy optimization for multimodal reasoning},
  author={Wang, Zhenhailong and Guo, Xuehang and Stoica, Sofia and Xu, Haiyang and Wang, Hongru and Ha, Hyeonjeong and Chen, Xiusi and Chen, Yangyi and Yan, Ming and Huang, Fei and others},
  journal={arXiv preprint arXiv:2507.06448},
  year={2025}
}

@inproceedings{zhang2024mathverse,
  title={Mathverse: Does your multi-modal llm truly see the diagrams in visual math problems?},
  author={Zhang, Renrui and Jiang, Dongzhi and Zhang, Yichi and Lin, Haokun and Guo, Ziyu and Qiu, Pengshuo and Zhou, Aojun and Lu, Pan and Chang, Kai-Wei and Qiao, Yu and others},
  booktitle={ECCV},
  pages={169--186},
  year={2024},
  organization={Springer}
}

@article{xiao2024logicvista,
  title={Logicvista: Multimodal llm logical reasoning benchmark in visual contexts},
  author={Xiao, Yijia and Sun, Edward and Liu, Tianyu and Wang, Wei},
  journal={arXiv preprint arXiv:2407.04973},
  year={2024}
}

@article{chen2024we,
  title={Are we on the right way for evaluating large vision-language models?},
  author={Chen, Lin and Li, Jinsong and Dong, Xiaoyi and Zhang, Pan and Zang, Yuhang and Chen, Zehui and Duan, Haodong and Wang, Jiaqi and Qiao, Yu and Lin, Dahua and others},
  journal={NeurIPS},
  volume={37},
  pages={27056--27087},
  year={2024}
}

@inproceedings{guan2024hallusionbench,
  title={Hallusionbench: an advanced diagnostic suite for entangled language hallucination and visual illusion in large vision-language models},
  author={Guan, Tianrui and Liu, Fuxiao and Wu, Xiyang and Xian, Ruiqi and Li, Zongxia and Liu, Xiaoyu and Wang, Xijun and Chen, Lichang and Huang, Furong and Yacoob, Yaser and others},
  booktitle={CVPR},
  pages={14375--14385},
  year={2024}
}

@misc{Singh_2025, 
    title={GPT-5 system card}, 
    url={https://openai.com/index/gpt-5-system-card/}, 
    journal={OpenAI}, 
    author={Singh, Aaditya and Fry, Adam and Perelman, Adam and others}, 
    year={2025}}

@article{qu2025optimizing,
  title={Optimizing test-time compute via meta reinforcement fine-tuning},
  author={Qu, Yuxiao and Yang, Matthew YR and Setlur, Amrith and Tunstall, Lewis and Beeching, Edward Emanuel and Salakhutdinov, Ruslan and Kumar, Aviral},
  journal={ICML},
  year={2025}
}

@article{yang2025look,
  title={Look-Back: Implicit Visual Re-focusing in MLLM Reasoning},
  author={Yang, Shuo and Niu, Yuwei and Liu, Yuyang and Ye, Yang and Lin, Bin and Yuan, Li},
  journal={arXiv preprint arXiv:2507.03019},
  year={2025}
}

@article{feng2025video,
  title={Video-r1: Reinforcing video reasoning in mllms},
  author={Feng, Kaituo and Gong, Kaixiong and Li, Bohao and Guo, Zonghao and Wang, Yibing and Peng, Tianshuo and Wu, Junfei and Zhang, Xiaoying and Wang, Benyou and Yue, Xiangyu},
  journal={arXiv preprint arXiv:2503.21776},
  year={2025}
}

@article{vl-rethinker,
      title={VL-Rethinker: Incentivizing Self-Reflection of Vision-Language Models with Reinforcement Learning},
      author = {Wang, Haozhe and Qu, Chao and Huang, Zuming and Chu, Wei and Lin,Fangzhen and Chen, Wenhu},
      journal={arXiv preprint arXiv:2504.08837},
      year={2025}
}

@article{bai2025qwen2,
  title={Qwen2. 5-vl technical report},
  author={Bai, Shuai and Chen, Keqin and Liu, Xuejing and Wang, Jialin and Ge, Wenbin and Song, Sibo and Dang, Kai and Wang, Peng and Wang, Shijie and Tang, Jun and others},
  journal={arXiv preprint arXiv:2502.13923},
  year={2025}
}

@article{lu2023mathvista,
  title={Mathvista: Evaluating mathematical reasoning of foundation models in visual contexts},
  author={Lu, Pan and Bansal, Hritik and Xia, Tony and Liu, Jiacheng and Li, Chunyuan and Hajishirzi, Hannaneh and Cheng, Hao and Chang, Kai-Wei and Galley, Michel and Gao, Jianfeng},
  journal={ICLR},
  year={2023}
}

@article{wang2024measuring,
  title={Measuring multimodal mathematical reasoning with math-vision dataset},
  author={Wang, Ke and Pan, Junting and Shi, Weikang and Lu, Zimu and Ren, Houxing and Zhou, Aojun and Zhan, Mingjie and Li, Hongsheng},
  journal={NeurIPS},
  volume={37},
  pages={95095--95169},
  year={2024}
}

@article{qiao2024we,
  title={We-math: Does your large multimodal model achieve human-like mathematical reasoning?},
  author={Qiao, Runqi and Tan, Qiuna and Dong, Guanting and Wu, Minhui and Sun, Chong and Song, Xiaoshuai and GongQue, Zhuoma and Lei, Shanglin and Wei, Zhe and Zhang, Miaoxuan and others},
  journal={ACL},
  year={2024}
}

@article{lu2021inter,
  title={Inter-gps: Interpretable geometry problem solving with formal language and symbolic reasoning},
  author={Lu, Pan and Gong, Ran and Jiang, Shibiao and Qiu, Liang and Huang, Siyuan and Liang, Xiaodan and Zhu, Song-Chun},
  journal={ACL},
  year={2021}
}

@inproceedings{yue2024mmmu,
  title={Mmmu: A massive multi-discipline multimodal understanding and reasoning benchmark for expert agi},
  author={Yue, Xiang and Ni, Yuansheng and Zhang, Kai and Zheng, Tianyu and Liu, Ruoqi and Zhang, Ge and Stevens, Samuel and Jiang, Dongfu and Ren, Weiming and Sun, Yuxuan and others},
  booktitle={CVPR},
  pages={9556--9567},
  year={2024}
}

@article{yu2023mm,
  title={Mm-vet: Evaluating large multimodal models for integrated capabilities},
  author={Yu, Weihao and Yang, Zhengyuan and Li, Linjie and Wang, Jianfeng and Lin, Kevin and Liu, Zicheng and Wang, Xinchao and Wang, Lijuan},
  journal={ICML},
  year={2023}
}

@article{comanici2025gemini,
  title={Gemini 2.5: Pushing the frontier with advanced reasoning, multimodality, long context, and next generation agentic capabilities},
  author={Comanici, Gheorghe and Bieber, Eric and Schaekermann, Mike and Pasupat, Ice and Sachdeva, Noveen and Dhillon, Inderjit and Blistein, Marcel and Ram, Ori and Zhang, Dan and Rosen, Evan and others},
  journal={arXiv preprint arXiv:2507.06261},
  year={2025}
}

@article{chen2024expanding,
  title={Expanding performance boundaries of open-source multimodal models with model, data, and test-time scaling},
  author={Chen, Zhe and Wang, Weiyun and Cao, Yue and Liu, Yangzhou and Gao, Zhangwei and Cui, Erfei and Zhu, Jinguo and Ye, Shenglong and Tian, Hao and Liu, Zhaoyang and others},
  journal={arXiv preprint arXiv:2412.05271},
  year={2024}
}

@inproceedings{duan2024vlmevalkit,
  title={Vlmevalkit: An open-source toolkit for evaluating large multi-modality models},
  author={Duan, Haodong and Yang, Junming and Qiao, Yuxuan and Fang, Xinyu and Chen, Lin and Liu, Yuan and Dong, Xiaoyi and Zang, Yuhang and Zhang, Pan and Wang, Jiaqi and others},
  booktitle={ACM MM},
  pages={11198--11201},
  year={2024}
}

@inproceedings{wang2025mv,
  title={Mv-math: Evaluating multimodal math reasoning in multi-visual contexts},
  author={Wang, Peijie and Li, Zhong-Zhi and Yin, Fei and Ran, Dekang and Liu, Cheng-Lin},
  booktitle={CVPR},
  pages={19541--19551},
  year={2025}
}

@article{chen2024m,
  title={M3CoT: A Novel Benchmark for Multi-Domain Multi-step Multi-modal Chain-of-Thought},
  author={Chen, Qiguang and Qin, Libo and Zhang, Jin and Chen, Zhi and Xu, Xiao and Che, Wanxiang},
  journal={ACL},
  year={2024}
}

@inproceedings{sheng2025hybridflow,
  title={Hybridflow: A flexible and efficient rlhf framework},
  author={Sheng, Guangming and Zhang, Chi and Ye, Zilingfeng and Wu, Xibin and Zhang, Wang and Zhang, Ru and Peng, Yanghua and Lin, Haibin and Wu, Chuan},
  booktitle={Proceedings of the Twentieth European Conference on Computer Systems},
  pages={1279--1297},
  year={2025}
}

@article{liu2023visual,
  title={Visual instruction tuning},
  author={Liu, Haotian and Li, Chunyuan and Wu, Qingyang and Lee, Yong Jae},
  journal={NeurIPS},
  volume={36},
  pages={34892--34916},
  year={2023}
}

@inproceedings{wu2024v,
  title={V?: Guided visual search as a core mechanism in multimodal llms},
  author={Wu, Penghao and Xie, Saining},
  booktitle={CVPR},
  pages={13084--13094},
  year={2024}
}

@article{wang2025think,
  title={Think or Not? Selective Reasoning via Reinforcement Learning for Vision-Language Models},
  author={Wang, Jiaqi and Lin, Kevin Qinghong and Cheng, James and Shou, Mike Zheng},
  journal={arXiv preprint arXiv:2505.16854},
  year={2025}
}

@article{zhang2025qwen3,
  title={Qwen3 Embedding: Advancing Text Embedding and Reranking Through Foundation Models},
  author={Zhang, Yanzhao and Li, Mingxin and Long, Dingkun and Zhang, Xin and Lin, Huan and Yang, Baosong and Xie, Pengjun and Yang, An and Liu, Dayiheng and Lin, Junyang and others},
  journal={arXiv preprint arXiv:2506.05176},
  year={2025}
}

@article{wang2024muirbench,
  title={Muirbench: A comprehensive benchmark for robust multi-image understanding},
  author={Wang, Fei and Fu, Xingyu and Huang, James Y and Li, Zekun and Liu, Qin and Liu, Xiaogeng and Ma, Mingyu Derek and Xu, Nan and Zhou, Wenxuan and Zhang, Kai and others},
  journal={arXiv preprint arXiv:2406.09411},
  year={2024}
}

@article{zou2025unlocking,
  title={Unlocking Vision-Language Models for Video Anomaly Detection via Fine-Grained Prompting},
  author={Zou, Shu and Tian, Xinyu and Wesemann, Lukas and Waschkowski, Fabian and Yang, Zhaoyuan and Zhang, Jing},
  journal={arXiv preprint arXiv:2510.02155},
  year={2025}
}

@article{yao2025simple,
  title={Simple radiology vllm test-time scaling with thought graph traversal},
  author={Yao, Yue and Wen, Zelin and Tong, Yan and Tian, Xinyu and Li, Xuqing and Ma, Xiao and Xu, Dongliang and Gedeon, Tom},
  journal={arXiv preprint arXiv:2506.11989},
  year={2025}
}

@article{he2025few,
  title={Few Tokens Matter: Entropy Guided Attacks on Vision-Language Models},
  author={He, Mengqi and Tian, Xinyu and Shen, Xin and Ni, Jinhong and Zou, Shu and Yang, Zhaoyuan and Zhang, Jing},
  journal={arXiv preprint arXiv:2512.21815},
  year={2025}
}

@inproceedings{tian2024argue,
  title={Argue: Attribute-guided prompt tuning for vision-language models},
  author={Tian, Xinyu and Zou, Shu and Yang, Zhaoyuan and Zhang, Jing},
  booktitle={Proceedings of the IEEE/CVF Conference on Computer Vision and Pattern Recognition},
  pages={28578--28587},
  year={2024}
}

@inproceedings{tian2025identifying,
  title={Identifying and mitigating position bias of multi-image vision-language models},
  author={Tian, Xinyu and Zou, Shu and Yang, Zhaoyuan and Zhang, Jing},
  booktitle={Proceedings of the Computer Vision and Pattern Recognition Conference},
  pages={10599--10609},
  year={2025}
}

@article{yao2023training,
  title={Training with product digital twins for autoretail checkout},
  author={Yao, Yue and Tian, Xinyu and Tang, Zheng and Biswas, Sujit and Lei, Huan and Gedeon, Tom and Zheng, Liang},
  journal={arXiv preprint arXiv:2308.09708},
  year={2023}
}

@article{tian2025black,
  title={Black sheep in the herd: Playing with spuriously correlated attributes for vision-language recognition},
  author={Tian, Xinyu and Zou, Shu and Yang, Zhaoyuan and He, Mengqi and Zhang, Jing},
  journal={arXiv preprint arXiv:2502.15809},
  year={2025}
}

@inproceedings{zou2025simlabel,
  title={Simlabel: Consistency-guided ood detection with pretrained vision-language models},
  author={Zou, Shu and Tian, Xinyu and Zhao, Qinyu and Yang, Zhaoyuan and Zhang, Jing},
  booktitle={Australasian Joint Conference on Artificial Intelligence},
  pages={110--121},
  year={2025},
  organization={Springer}
}
\bibliographystyle{iclr2026_conference}

\clearpage
\setcounter{secnumdepth}{0}
\setcounter{tocdepth}{4}
\renewcommand{\baselinestretch}{1.6}\normalsize
{\sc \tableofcontents}
\renewcommand{\baselinestretch}{1.0}\normalsize
\clearpage

\section{A \quad Experimental Details}
\subsection{A.1 \quad Benchmark Statistics}
In the main paper, we provide a high-level overview of the selected benchmarks. Here, we present detailed information for clarity and ease of reproducibility, including dataset types, sizes, and splits. All evaluation scripts are implemented using the VLMEvalKit framework~\citep{duan2024vlmevalkit}.

1) \textbf{MathVerse} is a benchmark comprising 2,612 high-quality mathematical questions. Each example provides varying levels of multimodal information. Following prior work, we evaluate on the Test Mini split using the Vision Only setting, which includes approximately 700 samples.

2) \textbf{MathVista} is a comprehensive mathematical benchmark that evaluates a range of skills including puzzle solving, algebraic reasoning, and scientific understanding, and comprises 6,141 examples. For our evaluation, we use the Test Mini split, which contains approximately 1,000 examples.

3) \textbf{MathVision} comprises 3,040 high-quality mathematical problems sourced from real-world math competitions, spanning 16 distinct disciplines and five levels of difficulty, providing a comprehensive and challenging benchmark for evaluating VLMs. We conduct our evaluation using the full test set.

4) \textbf{LogicVista} assesses the fundamental logical reasoning capabilities of VLMs, covering a range of reasoning types including spatial, deductive, inductive, numeric, and mechanical. The benchmark comprises 448 visual multiple-choice questions. We conduct our evaluation on the full test set.

5) \textbf{WeMath} comprises 6.5k visual math questions structured around 67 hierarchical knowledge concepts across five levels of granularity. We evaluate on the Test Mini split, which contains approximately 1,740 examples, and report the strict score as the primary evaluation metric.

6) \textbf{Geometry3k} consists of 3,002 geometry problems with dense annotations in formal language, requiring abstract problem-solving and symbolic reasoning based on axiomatic knowledge. For evaluation, we combine the validation and test splits, resulting in approximately 900 examples.

7) \textbf{MMMU} is a challenging benchmark that covers a broad range of disciplines, requiring college-level subject knowledge and reasoning. It contains 11.5k curated multimodal questions sourced from college exams, quizzes, and textbooks. We perform our evaluation on the validation split.

8) \textbf{MMStar} is a vision-indispensable multimodal benchmark specifically designed to ensure that each sample exhibits strong visual dependency and requires advanced multimodal reasoning capabilities. It comprises 1,500 samples for offline evaluation. Here we evaluate on the full test set.

9) \textbf{HallusionBench} is designed to challenge advanced VLMs by emphasizing fine-grained understanding and interpretation of visual information. It consists of 346 manually curated images paired with 1,129 questions. In this study, we conduct our evaluation on the full test split.

10) \textbf{MMVet} comprises 218 examples and defines six core vision-language capabilities, focusing on their integration to evaluate the synergy among different skills. It is designed to assess the overall competence of generalist models. We perform our evaluation on the full test split.

\subsection{A.2 \quad Baseline Settings}
In the main text, we compare our method against several popular baselines to validate its effectiveness. For models such as GPT-5-Thinking, Gemini-2.5-Pro, and InternVL-2.5, we directly report results from the OpenCompass leaderboard. For other models, we provide reproduced results when official numbers are not available in the original papers. Below, we detail the settings used for these reproduced baseline models for reference. All results are reproduced using greedy decoding.

1) \textbf{R1-OneVision} includes both 3B and 7B variants, which are first trained via SFT followed by RL, with a particular focus on mathematical reasoning tasks. In our experiments, we adopt the publicly available 7B checkpoint based on Qwen2.5-VL-7B.

2) \textbf{VLAA-Thinker} is an RL-only model trained on a high-quality and challenging dataset, and is the first to demonstrate that RL outperforms SFT in multimodal settings. In our experiments, we use the VLAA-Thinker-7B variant, which is also based on Qwen2.5-VL-7B.

3) \textbf{Vision-R1} is a powerful reasoning model trained at scale, with the combined amount of RL and SFT data being approximately five times larger than that used in our setting. In our experiments, we use Vision-R1-7B, which represents the strongest variant reported by the authors.

\subsection{A.3 \quad Error Analysis}
\label{subsec:A3}

\begin{figure*}[t!] 
    \centering
    \includegraphics[width=1\linewidth]{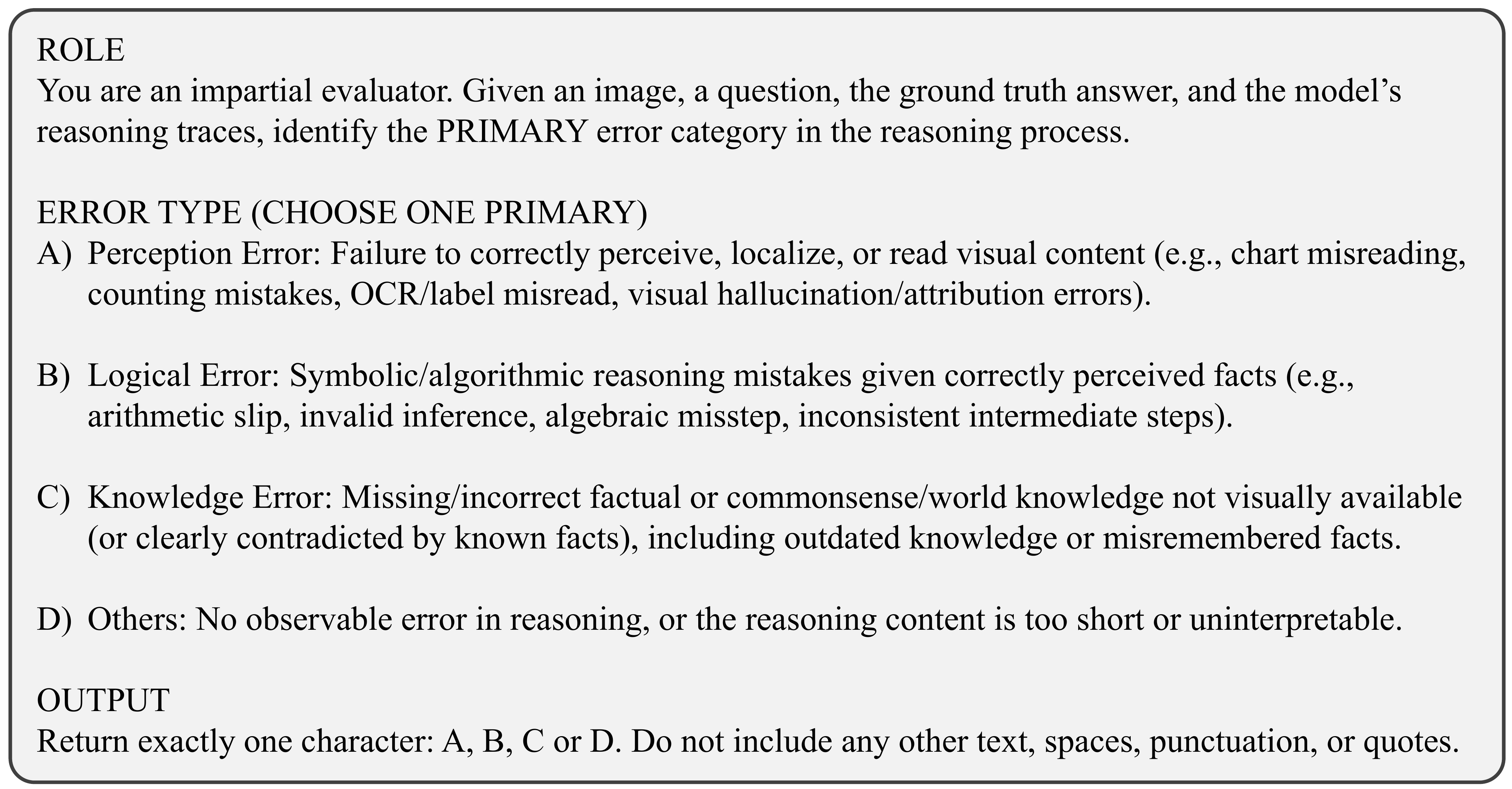}
    \caption{The designed prompt to instruct GPT-5 to assist in categorizing the different error types.
    }
\label{fig:error_prompt}
\end{figure*}

\begin{figure*}[b!] 
    \centering
    \includegraphics[width=1\linewidth]{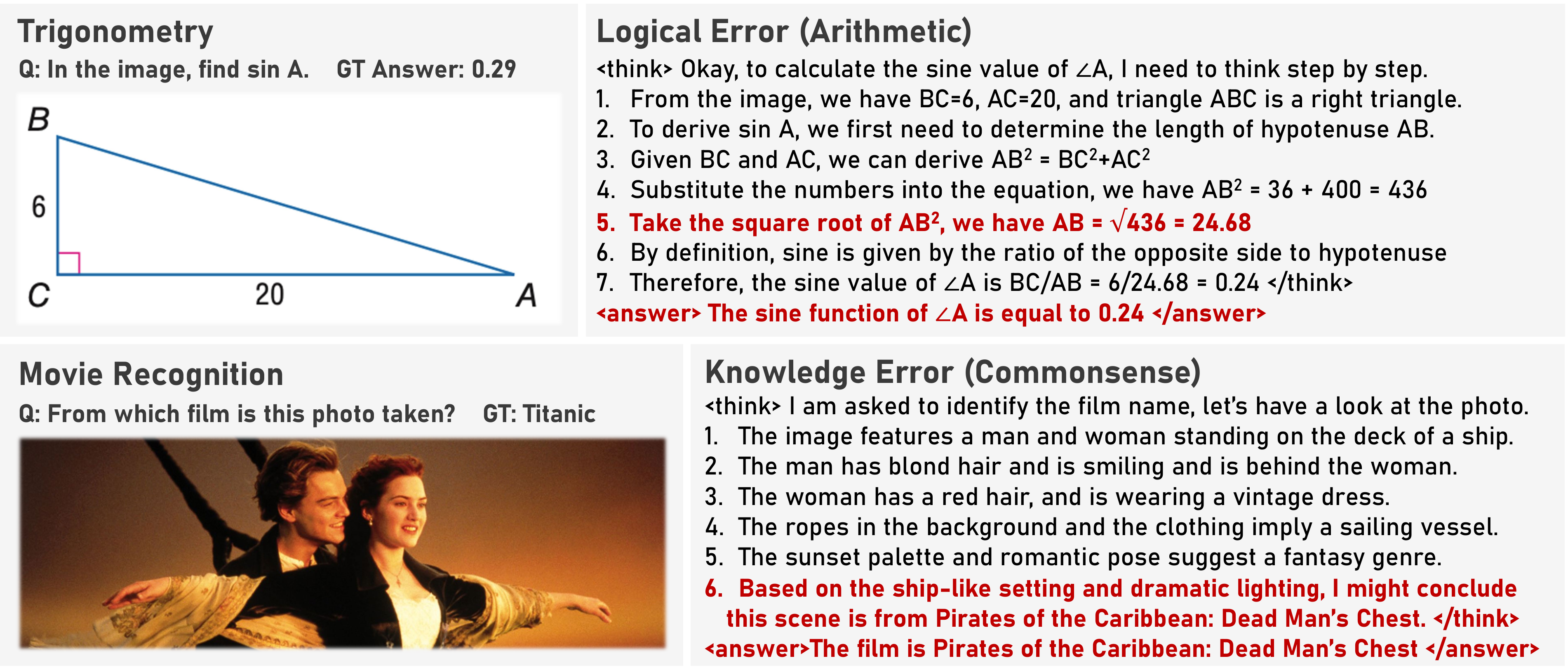}
    \caption{The representative examples of logical and knowledge errors.
    }
\label{fig:error_example}
\end{figure*}

In Section~\hyperref[subsec:3.1]{3.1}, we conduct a comprehensive error analysis by leveraging GPT-5 to categorize different types of errors and quantify their distributions, aiming to assess the impact of reasoning on failure cases. The prompt used for GPT-5 input is shown in Fig.~\ref{fig:error_prompt}. As illustrated, we provide GPT-5 with four options: three predefined error categories, along with an Others category to account for a small number of outliers, such as cases with uninterpretable content. We then compute the proportion of each error type to support our analysis. For completeness, in addition to the perception error examples as shown in Fig.~\ref{fig:motivation}, here we also provide representative cases of logical errors and knowledge errors, as illustrated in Fig.~\ref{fig:error_example}.

\begin{table*}[t]
\footnotesize
\centering
\setlength{\tabcolsep}{1.6mm}
\setlength\heavyrulewidth{0.2ex}
\renewcommand{\arraystretch}{1.0}
\begin{NiceTabular}{@{}c|cccc@{}}
\toprule
Error Category & Logic $\rightarrow$ Perc & Perc $\rightarrow$ Logic & Knowledge $\rightarrow$ Perc & Knowledge $\rightarrow$ Logic \\ \midrule
\# Examples    & 5                                & 7                                & 3                                  & 1                               \\ \bottomrule
\end{NiceTabular}
\caption{The number of examples with multiple error categories co-occuring.}
\label{tab:co-occur}
\end{table*}

\textcolor{blue}{In addition, based on the recorded human audit of $500$ randomly sampled failure cases, we find that instances containing multiple error types within the same response are quite rare, occurring in fewer than $20$ examples ($<4\%$), which is also the reason we prompt GPT to directly identify the primary error category as shown in Fig.~\ref{fig:error_prompt}. To further investigate this, we manually re-examine these cases and record the specific ordering patterns in which different error categories co-occur. As shown in Table~\ref{tab:co-occur}, any error category combinations not listed above did not occur in our samples. Notably, cases in which logical errors precede perceptual errors are extremely rare. This rarity ensures that such patterns do not meaningfully affect our experimental results or conclusions. It also makes it difficult to draw any substantive connection between perceptual errors and other error types, given how infrequently they co-occur in practice. These results further reinforce our finding that longer reasoning contexts tend to induce visual forgetting, which in turn leads to the observed performance degradation.}

\subsection{A.4 \quad Attention Visualization}
\label{subsec:A4}

In the main paper, we visualize the attention over image tokens to reflect the contribution of visual information to the model's reasoning process. Here, we provide additional details on the implementation. For each generation step, we compute the sum of attention scores after softmax assigned by the output token to all preceding image tokens. This yields a ratio between 0 and 1, which is then averaged across all layers to produce the final attention value. Additionally, in Fig.~\ref{fig:forget}~(A), since inserting images, \ie, visual replay, or instructions, \ie, focus prompt, may alter the final reasoning trajectory and result in slightly different sequence lengths, we normalize the comparison by truncating all outputs to the first 250 tokens, which basically covers the full response. This strategy is also applied in Fig.~\ref{fig:result}~(A) when comparing the visual attention between Vision-R1 and \texttt{VAPO-Thinker}.

\begin{figure*}[b!] 
    \centering
    \includegraphics[width=1\linewidth]{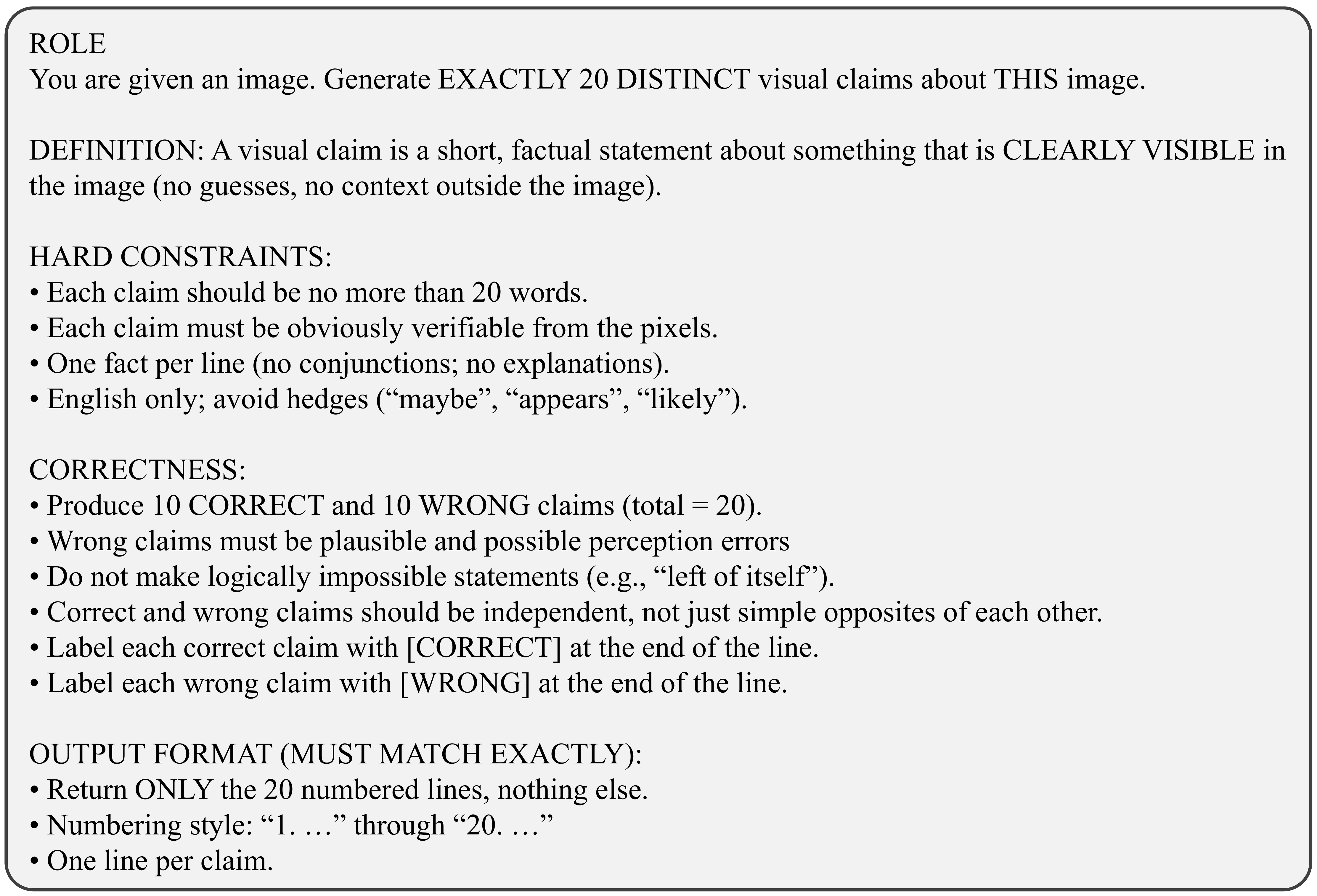}
    \caption{The user prompt for GPT-5 to generate visual claims for \texttt{VAPO}.
    }
\label{fig:claim_prompt}
\end{figure*}

\subsection{A.5 \quad Inference-level Remedies}
\label{subsec:A5}

In the main text, we introduce two inference-level remedies, \ie, visual replay and focus prompt, as preliminary strategies to demonstrate the negative impact of visual forgetting on reasoning. For completeness, we provide a detailed description of both approaches here.

In the visual replay strategy, the input image is reintroduced periodically during the model's reasoning process at regular intervals. In practice, to improve efficiency and prevent exceeding the model's context length, the reinserted image is downsampled to a lower resolution. Furthermore, the insertion points are selected to satisfy two criteria: (1) uniform segmentation of the reasoning trajectory, and (2) alignment with logical boundaries such as commas, periods, or line breaks to avoid interrupting syntactic or semantic units.

For the focus prompt strategy, at each insertion point, we randomly sample one prompt from a set of three manually designed instructions including \enquote{$\rm I \ need \ to \ see \ the \ image$}, \enquote{$\rm I \ have \ to \ look \ back$} and \enquote{$\rm Let \ me \ verify \ against \ the \ visual \ input$}, to ensure robustness against prompt variability. For alignment, we adopt the nearby insertion positions as used in visual replay, facilitating a consistent comparison between the two methods. It is worth noting that in visual replay, the image is reintroduced as part of the user prompt in a dialogue format, whereas in focus prompt, the instruction is directly injected into the assistant’s response.

\subsection{A.6 \quad Visual Claim Generation}
\label{subsec:A6}

\begin{figure*}[b!] 
    \centering
    \includegraphics[width=1\linewidth]{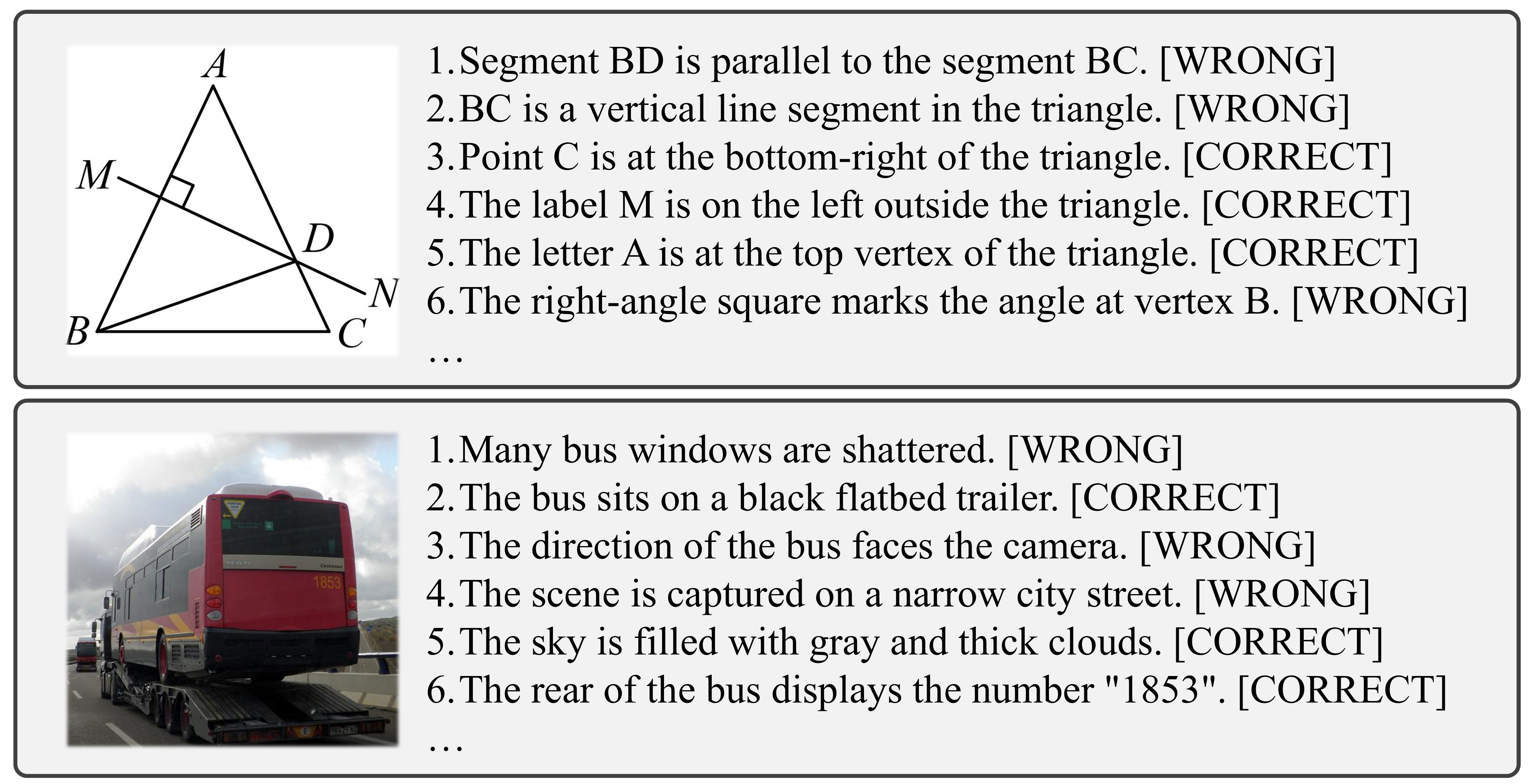}
    \caption{The examples of the GPT-generated visual claims.
    }
\label{fig:claim_example}
\end{figure*}

Our method encourages the model to rely more effectively on visual input by introducing a perception reward, which is derived from the model’s ability to evaluate a set of visual claims during the reasoning process. This evaluation serves as a proxy for assessing the model’s perceptual capability at varying reasoning stages. To support this, we prompt GPT-5 to generate a specified number of visual claims conditioned on the input image. An example of the user prompt used for this purpose is shown in Fig.~\ref{fig:claim_prompt}. We require the generated visual claims to be concise factual statements, free from hedging terms such as $\rm possibly$ or $\rm appears$. Each set of claims must contain an equal number of correct and incorrect statements. Furthermore, the claims are constructed to be independent of the corresponding example question, ensuring that the model must refer to the visual input rather than relying on prior outputs when evaluating their validity. While several generated visual claims have already been presented in Fig.~\ref{fig:method}, we provide more illustrative examples in Fig.~\ref{fig:claim_example} for reference.

\textcolor{blue}{In addition, we employ a rule-based filtering mechanism and dynamically generate visual claims. That is, for each training example, any claim that fails to meet the required criteria is discarded and resampled until the target number of valid claims is reached. The filtering process incorporates multiple constraints, as detailed in the prompt specification in Fig.~\ref{fig:claim_prompt}, including limits on claim length, the prohibition of ambiguous expressions (e.g., ``${\rm maybe}$'', ``${\rm appears}$''), and the requirement to maintain a balanced ratio of correct and incorrect claims. To reduce redundancy, we compute pairwise embedding distances between generated claims using Qwen3-Embedding-0.6B~\citep{zhang2025qwen3}, and remove claims with excessively high semantic similarity ($> 0.95$). We allow straightforward low-level claims such as color naming, while we impose a mild diversity constraint that limits the number of purely low-level color or counting statements per image through regular expression, favoring a richer distribution of relational, text-based, and global scene-level claims. }

\subsection{A.7 \quad Training Configuration}
\label{subsec:A7}

In Section 5, we briefly outline the training setup. Here, we provide a more detailed description. By default, we adopt ViRL39K as our training dataset, which is a refined collection derived from multiple existing sources such as MM-Eureka~\citep{meng2025mm}, MV-Math~\citep{wang2025mv}, and M3CoT~\citep{chen2024m}. The dataset covers a wide spectrum of domains, including STEM subjects, social topics, chart reasoning, and spatial relations. It is worth noting that prior baselines rely on substantially larger datasets that combine SFT and RL, for example, 155k samples for R1-OneVision and 210k samples for Vision-R1. Although this comparison is not fair for us in terms of training data scale, it further highlights the effectiveness of our proposed method.

During the rollout phase, we sample $5$ responses per example with a temperature of $1.0$, and employ vLLM as the backend to accelerate decoding. We then insert visual anchors into these generated responses to evaluate the model’s perceptual capability. For each anchor, we randomly sample a prefix of the reasoning content, append a visual claim, and instruct the model to judge its correctness. To ensure binary decision-making, the decoding process is constrained to produce either $\rm yes$ or $\rm no$, with the temperature fixed at $0.0$. Importantly, throughout this stage, we adopt the default system prompt of Qwen2.5-VL without explicitly introducing the presence or meaning of anchors. This design avoids altering the model’s inherent behavior and keeps anchors fully transparent, serving as an implicit reward mechanism that encourages the model to rely more heavily on visual cues. 

For training, we employ 8 NVIDIA A100-80G GPUs. The policy loss follows the default configuration of GRPO, where the clip ratio $\epsilon$ is set to $0.2$, and the KL penalty coefficient $\lambda$ is fixed at $1e^{-2}$. For \texttt{VAPO}, we set the anchor number $K=20$, the late-emphasis weight $\beta = 1.5$, and the perception reward weight $\gamma = 0.1$. The entire training process is conducted using verl~\citep{sheng2025hybridflow}.

\section{B \quad More Evaluation}
\label{sec:B}

\subsection{B.1 \quad Error Category Distribution}

\begin{figure*}[b!] 
    \centering
    \includegraphics[width=0.9\linewidth]{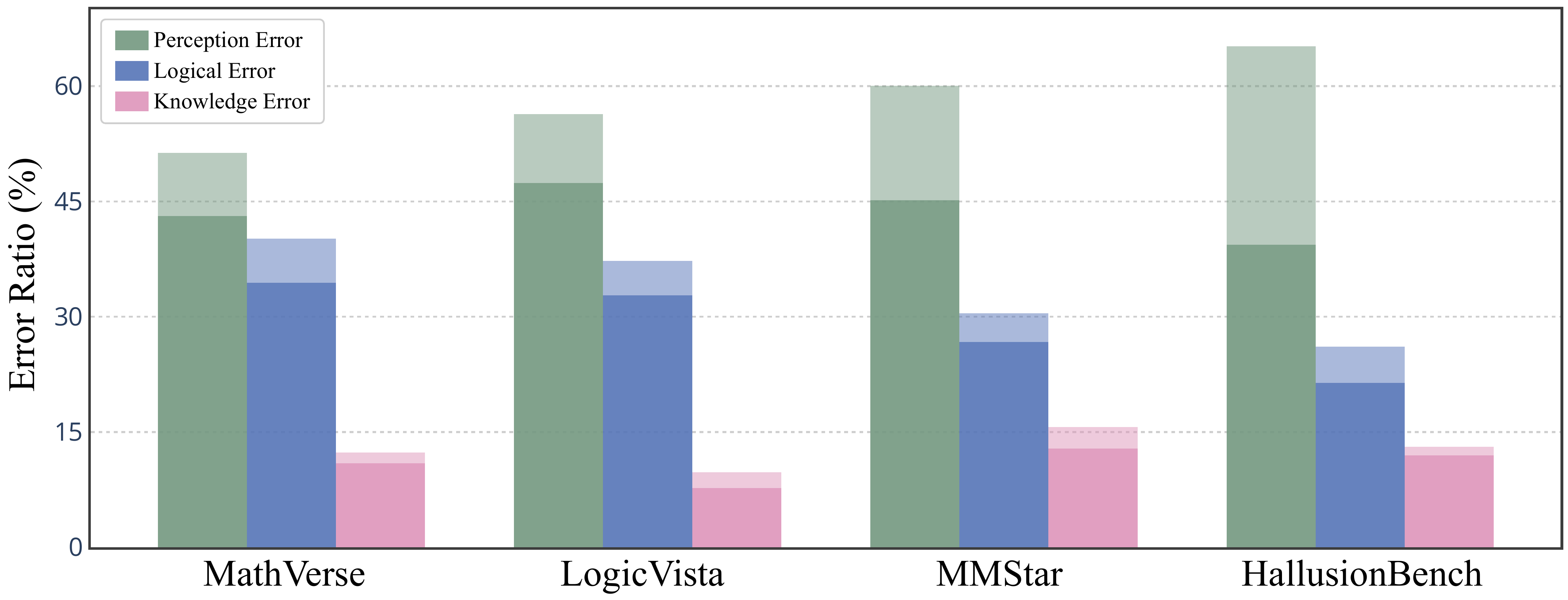}
    \caption{The error ratio of Vision-R1 as well as the correction rate achieved by our method across benchmarks. In the bar chart, the full height of each bar represents the overall error rate of Vision-R1, with the light-colored segment indicating the proportion of errors successfully corrected by our method, and the dark-colored segment corresponding to the remaining uncorrected errors.
    }
\label{fig:error_correct}
\end{figure*}

In the main text, we analyze the error cases of the representative baseline Vision-R1 and observe that perception errors constitute the majority, underscoring the detrimental impact of reasoning on visual information utilization. To assess the effectiveness of our approach in addressing this issue, we further examine the correction rate of our method across the different error categories of the baseline model, as illustrated in Fig.~\ref{fig:error_correct}. We observe that, for cases where the baseline model fails, our method substantially corrects a significant portion of perception errors, with improvements reaching up to $20\%$ on vision-intensive benchmarks. This perceptual correction capability constitutes a major source of the performance gains achieved by our approach.

\begin{figure*}[b!] 
    \centering
    \includegraphics[width=0.8\linewidth]{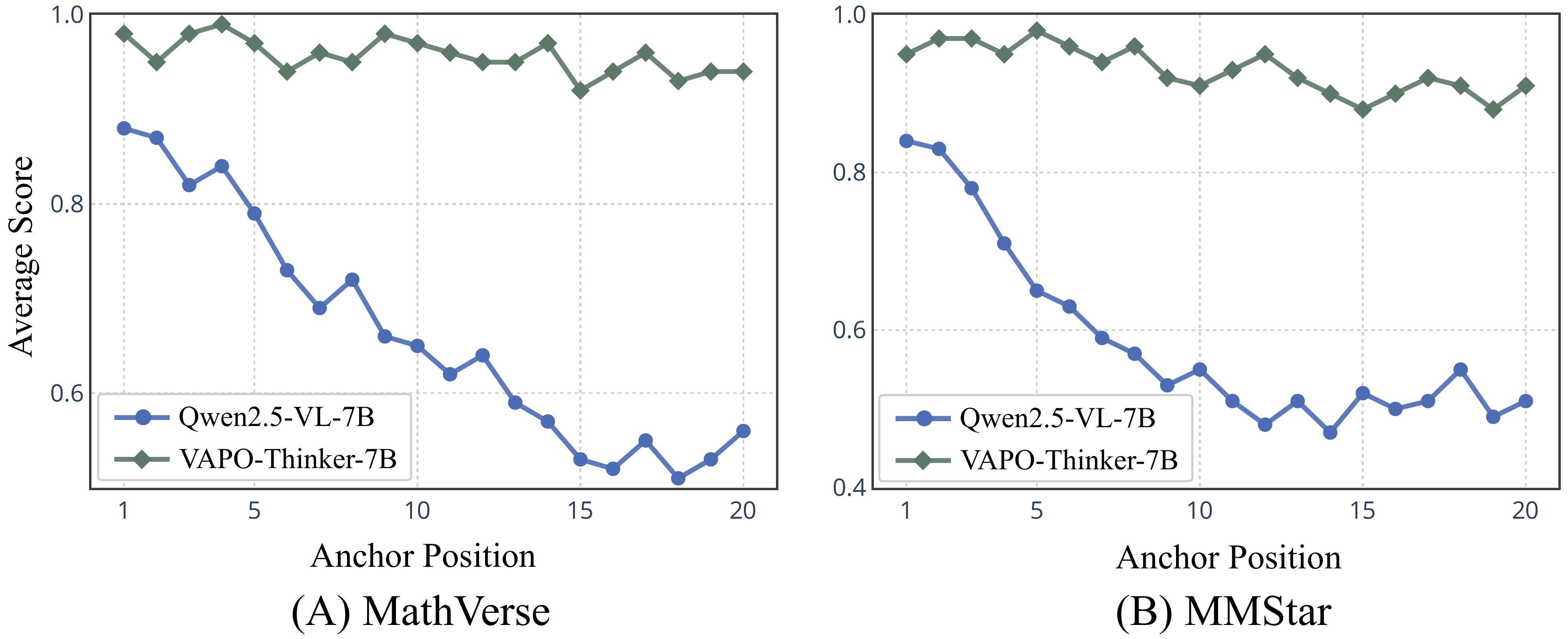}
    \caption{The anchor score across varying anchor positions, where a higher index corresponds to a later stage in the reasoning process. Scores are averaged across all examples within the benchmark.
    }
\label{fig:anchor_score}
\end{figure*}

\subsection{B.2 \quad Visual Anchor Score}

Our method assesses the model’s overall perceptual capability by setting up visual anchors at different stages of the reasoning process. As shown in Fig.~\ref{fig:dynamic}, the perception reward exhibits a clear upward trend throughout training, indicating that the model becomes increasingly proficient at evaluating visual claims and progressively relies more on visual information. To further investigate the role of individual visual anchors, we visualize the distribution of anchor scores before and after training, \ie, comparing the base model with the model after applying our method. As shown in Fig.~\ref{fig:anchor_score}, the base model, \ie, Qwen2.5-VL-7B, exhibits a clear downward trend in anchor scores as reasoning progresses, eventually approaching the level of random guessing (around $0.5$). In contrast, our trained model significantly improves anchor scores, maintaining a stable level around $0.9$ throughout the reasoning process. This indicates a substantial enhancement in perceptual capability as a result of our method.

\subsection{B.3 \quad Perception Reward Weight}
\label{subsec:B3}

\begin{table*}[t]
\footnotesize
\centering
\setlength{\tabcolsep}{1.6mm}
\setlength\heavyrulewidth{0.2ex}
\renewcommand{\arraystretch}{1.0}
\begin{tabular}{@{}cccccccc@{}}
\toprule
$\gamma$ & MathVerse     & LogicVista    & Geometry3k & MMMU & MMStar        & HallBench & Avg. \\ \midrule
0.00     & 44.2          & 44.4          & 44.5                           & 53.6                     & 55.2          & 51.2                          & 49.0                     \\
0.05  & 47.7          & 46.3          & \textbf{48.2}                  & 55.5                     & 58.6          & 53.7                          & 51.7                     \\
0.10   & \textbf{48.9} & 47.3          & 47.7                           & \textbf{56.7}            & 59.1          & 55.5                          & \textbf{52.5}            \\
0.15  & 48.4          & \textbf{46.7} & 48.1                           & 56.2                     & \textbf{59.6} & \textbf{56.7}                 & 52.4                     \\
0.20   & 47.9          & 46.6          & 47.4                           & 55.7                     & 59.4          & 54.2                          & 51.9                     \\ \bottomrule
\end{tabular}
\caption{The benchmark accuracy with varying perception reward weight $\gamma$.}
\label{tab:gamma}
\end{table*}

In the main text, we have conducted ablation studies on the anchor number $K$ and the late-emphasis weight $\beta$ to identify their optimal settings. Here, we further investigate the impact of perception reward weight $\gamma$ on model performance. For computational efficiency, in this experiment we do not train on the full dataset but instead randomly sample $5000$ examples from the full $39$k training set. As shown in Table~\ref{tab:gamma}, setting $\gamma = 0$ reduces the training procedure to vanilla GRPO without perceptual supervision. As $\gamma$ increases, the average accuracy across benchmarks improves significantly, peaking around $\gamma = 0.1$. Notably, mathematical tasks tend to favor smaller values of $\gamma$, which is intuitive given the relatively simple visual structures of these tasks. In contrast, vision-intensive benchmarks such as MMStar and HallusionBench benefit from more aggressive settings, reflecting their greater reliance on strong perceptual capability.

\subsection{B.4 \quad Ablation on Data Augmentation}

\begin{table*}[t]
\footnotesize
\centering
\setlength{\tabcolsep}{1.6mm}
\setlength\heavyrulewidth{0.2ex}
\renewcommand{\arraystretch}{1.0}
\begin{NiceTabular}{@{}lccccccc@{}}
\toprule
\multicolumn{1}{c}{Method} & MathVerse & LogicVista & Geometry3k & MMMU & MMStar & HallBench & Avg. \\ \midrule
GRPO                       & 44.2      & 44.4       & 44.5       & 53.6 & 55.2   & 52.2      & 49.0 \\
GRPO\textsubscript{aug}                  & 44.7      & 45.2       & 43.9       & 54.2 & 54.8   & 53.1      & 49.3 \\
\rowcolor[HTML]{EDEDED} \texttt{VAPO}                       & \textbf{45.9}      & \textbf{46.1}       & \textbf{46.7}       & \textbf{55.2} & \textbf{57.8}   & \textbf{54.3 }     & \textbf{51.0} \\ \bottomrule
\end{NiceTabular}
\caption{The comparison with \texttt{VAPO} and GRPO augmented with visual claim examples.}
\label{tab:augmentation}
\end{table*}

In our approach, visual claims play a central role by providing reward signals that explicitly encourage the model to rely more heavily on visual inputs. This mechanism helps mitigate visual forgetting and improves accuracy across benchmarks. However, since these visual claims are generated by GPT and accompanied by correctness labels, they may be viewed as new synthetic examples augmented to the original training data. This raises a critical question: are the performance gains brought by \texttt{VAPO} primarily attributable to the reward-driven promotion of visually grounded reasoning, or are they simply a result of data augmentation through the inclusion of these synthetic visual examples?

To investigate this question, we introduce a new baseline in which the generated visual claims are directly transformed into additional training examples and integrated into the original dataset. For computational efficiency, we randomly sample $5000$ examples from the full dataset, as described in Appendix~\hyperref[subsec:B3]{B.3}. For each example, we generate $5$ visual claims, resulting in a fivefold expansion of the training set to $30$k examples. This setup allows us to isolate the performance gains attributable purely to data augmentation with synthetic visual claims. The baseline is trained using the standard GRPO algorithm without any additional modifications. In contrast, under the \texttt{VAPO} framework, these same visual claims are used as anchors during training, with the anchor number set to $K=5$ to match the available number of claims per instance.

As shown in Table~\ref{tab:augmentation}, augmenting vanilla GRPO with additional examples constructed from visual claims yields only marginal improvement over the original dataset ($49.0 \rightarrow49.3$). In contrast, our method achieves a substantial performance gain of $2\%$ ($49.0 \rightarrow 51.0$). This discrepancy may be attributed to two key factors: 1) The visual claims are generally short, simple, and strongly dependent on visual information, often solvable without long reasoning trajectories, thus offering limited learning capacity; 2) The augmented examples are derived from a small number of unique images, \ie, five examples share the same image, significantly reducing data diversity and increasing the risk of overfitting. These results suggest that interpreting the role of visual claims purely as a form of data augmentation is not appropriate. Rather, the performance gains of \texttt{VAPO} are primarily driven by its promotion of visually grounded reasoning.

\subsection{B.5 \quad Effect of Visual Claims}

\begin{table*}[t]
\footnotesize
\centering
\setlength{\tabcolsep}{1.6mm}
\setlength\heavyrulewidth{0.2ex}
\renewcommand{\arraystretch}{1.0}
\begin{NiceTabular}{@{}lccccccc@{}}
\toprule
\multicolumn{1}{c}{Method} & MathVerse     & LogicVista    & Geometry3k    & MMMU          & MMStar        & HallBench     & Avg.          \\ \midrule
GRPO                       & 44.2          & 44.4          & 44.5          & 53.6          & 55.2          & 51.2          & 49.0          \\
NonVisualClaim             & 45.7          & 44.9          & 44.3          & 54.8          & 57.2          & 51.9          & 49.8          \\
QAlignedClaim              & 44.9          & 45.8          & 46.1          & 55.3          & 57.8          & 53.5          & 50.6          \\
\rowcolor[HTML]{EDEDED} VisClaim                   & \textbf{48.9} & \textbf{47.3} & \textbf{47.7} & \textbf{56.7} & \textbf{59.1} & \textbf{55.5} & \textbf{52.5} \\ \bottomrule
\end{NiceTabular}
\caption{The comparison with non-visually-dependent claims and question-aligned claims.}
\label{tab:claim}
\end{table*}

\begin{table*}[t]
\footnotesize
\centering
\setlength{\tabcolsep}{1.6mm}
\setlength\heavyrulewidth{0.2ex}
\renewcommand{\arraystretch}{1.0}
\begin{tabular}{@{}cccccccc@{}}
\toprule
$\lambda$ ($1e^{-2}$) & MathVerse & LogicVista & Geometry3k & MMMU & MMStar & HallBench & Avg. \\ \midrule
0      & 48.1      & 47.1       & 47.3       & 57.1 & 59.3   & 55.3      & 52.4 \\
1      & \textbf{48.9}      & \textbf{47.3}       & 47.7       & 56.7 & 59.1   & \textbf{55.5}      & \textbf{52.5} \\
2      & 48.3      & 47.0       & 47.9       & \textbf{57.2} & 58.4   & 54.9      & 52.3 \\
5      & 48.1      & 46.2       & \textbf{48.3}       & 57.1 & \textbf{59.8}   & 54.3      & 52.3 \\ \bottomrule
\end{tabular}
\caption{The impact of KL penalty coefficient $\lambda$ to the benchmark accuracy.}
\label{tab:lambda}
\end{table*}

In our method, visual claims serve as a critical proxy for measuring a model’s perceptual capability. To better understand their influence, we investigate how different types of visual claims affect model performance. In addition to our default setting, we consider two alternative variants: 1) non-visually-dependent claims: although these are factual statements about the image, they emphasize external knowledge or logical reasoning rather than concrete visual details; 2) question-aligned claims: these are closely related to the specific question associated with the image, potentially allowing the model to assess claim correctness based on its own historical reasoning outputs rather than pure visual grounding. To generate these claims, we prompt GPT accordingly: for the first variant, we explicitly instruct GPT to produce claims that are not visually dependent; for the second, we provide both the image and the corresponding question, asking it to output claims relevant to the question content. For efficiency, we conduct the analysis using 5000 examples sampled from the full training set.

As shown in Table~\ref{tab:claim}, both the non-visually-dependent and question-aligned variants perform substantially worse than our default visual claim setup. This is likely due to the fact that non-visually-dependent claims can often be judged correctly without requiring strong perceptual capability, thereby diminishing the core purpose of \texttt{VAPO}. On the other hand, question-aligned claims are highly correlated with the question content, allowing the model to infer their correctness from its own reasoning traces without relying on visual input. These results highlight that the design of visual claims is crucial to the effectiveness of the \texttt{VAPO} framework.

\subsection{B.6 \quad KL Penalty Coefficient}

In the previous sections, we have analyzed the impact of \texttt{VAPO}-specific hyperparameters, including the anchor number $K$, the late-emphasis weight $\beta$, and the perception reward weight $\gamma$. For the underlying GRPO framework, we adopt the default setting for the KL penalty coefficient $\lambda = 1e^{-2}$. For completeness, here we also examine the effect of varying $\lambda$ on the performance of \texttt{VAPO}. To maintain computational efficiency, we conduct this analysis on a randomly sampled subset of 5000 examples from the full training set. As shown in Table~\ref{tab:lambda}, \texttt{VAPO} exhibits minimal sensitivity to the choice of $\lambda$. Regardless of whether the KL penalty is applied or varied in magnitude, the average accuracy fluctuates by no more than $0.2\%$.

\subsection{B.7 \quad VAPO with Inference-Level Remedies}

\begin{table*}[t]
\footnotesize
\centering
\setlength{\tabcolsep}{1.6mm}
\setlength\heavyrulewidth{0.2ex}
\renewcommand{\arraystretch}{1.0}
\begin{tabular}{@{}lccccccc@{}}
\toprule
\multicolumn{1}{c}{Method} & MathVerse & LogicVista & Geometry3k & MMMU & MMStar & HallBench & Avg. \\ \midrule
\texttt{VAPO}                       & 53.3      & 50.9       & 51.3       & 60.2 & 63.0   & 57.4      & 56.0 \\
\texttt{VAPO}\textsubscript{FP}                   & 53.1      & 51.2       & 50.7       & 60.5 & 63.4   & 57.9      & 56.1 \\
\texttt{VAPO}\textsubscript{VR}                   & 53.7      & 50.4       & 51.8       & 57.8 & 62.7   & 58.2      & 55.8 \\ \bottomrule
\end{tabular}
\caption{The impact of \texttt{VAPO} augmented with inference-level remedies.}
\label{tab:vapo_augment}
\end{table*}

To further compare the effectiveness of our method with inference-level remedies, \ie, visual replay and focus prompt, we augment \texttt{VAPO} by incorporating these two straightforward approaches. We follow the same insertion strategy as in Fig.~\ref{fig:forget}, where interventions are applied at four points throughout the reasoning process. As shown in Table~\ref{tab:vapo_augment}, augmenting our model with inference-level remedies for visual forgetting results in minimal impact, with performance fluctuations within approximately $0.2\%$. This suggests that our method has already substantially enhanced the model's reliance on visual input, effectively subsuming the benefits provided by these additional strategies.

\subsection{B.8 \quad Attention-Based Reward}

\begin{table*}[t]
\footnotesize
\centering
\setlength{\tabcolsep}{1.6mm}
\setlength\heavyrulewidth{0.2ex}
\renewcommand{\arraystretch}{1.0}
\begin{NiceTabular}{@{}lccccccc@{}}
\toprule
Method                   & MathVerse & LogicVista & Geometry3k & MMMU & MMStar & HallBench & Avg. \\ \midrule
GRPO & 48.2      & 45.5       & 47.3       & 56.6 & 58.9   & 53.2      & 51.6 \\
\texttt{VAPO}\textsubscript{attn}               & 47.6      & 45.8       & 46.4       & 57.3 & 59.4   & 54.8      & 51.8 \\
\rowcolor[HTML]{EDEDED} \texttt{VAPO}\textsubscript{perc}               & \textbf{53.3}      & \textbf{50.9}       & \textbf{51.3}       & \textbf{60.2} & \textbf{63.0}   & \textbf{57.4}      & \textbf{56.0} \\ \bottomrule
\end{NiceTabular}
\caption{The comparison with attention reward which directly maximizes visual attention ratio.}
\label{tab:attention}
\end{table*}

In the main text, we use the evolution of visual attention throughout the reasoning process as a proxy to examine whether our method mitigates visual forgetting. Therefore, a natural baseline to consider is a more trivial alternative: directly using the attention scores of image tokens as a reward signal to guide training. To explore this possibility, we introduce an attention reward, which quantifies the model’s overall visual attention ratio across layers during the reasoning process, as detailed in Appendix~\hyperref[subsec:A4]{A.4}. We replace the perception reward with this attention-based reward while keeping all other experimental configurations consistent with the main setup, termed as \texttt{VAPO}\textsubscript{attn}, whereas our original method using perception reward is referred to as \texttt{VAPO}\textsubscript{perc}.

As shown in Table~\ref{tab:attention}, naively maximizing visual attention provides little to no performance gain and even leads to noticeable degradation on certain tasks. We hypothesize that this is due to two key factors: 1) Although visual attention can serve as an indicator of the contribution of visual inputs to the model’s decision-making, it does not directly reflect whether the model is effectively utilizing visual features. Blindly encouraging higher attention may disrupt the base model’s learned distribution; 2) A higher visual attention ratio is not inherently better. As observed in earlier experiments, its values typically lie between $0$ and $0.3$. Unconstrained maximization of this ratio may lead to instability or training collapse in later stages.

\subsection{B.9 \quad Computational Efficency and Cost}

\begin{table*}[t]
\footnotesize
\centering
\setlength{\tabcolsep}{1.6mm}
\setlength\heavyrulewidth{0.2ex}
\renewcommand{\arraystretch}{1.0}
\begin{NiceTabular}{@{}ccccc@{}}
\toprule
Method & Epoch & Time   & Accuracy       & Gain  \\ \midrule
GRPO   & 2     & 18h46m & 51.54          &   \NA    \\
DAPO   & 2     & 25h11m & 53.17          & +1.63 \\
\rowcolor[HTML]{EDEDED} \texttt{VAPO}   & 2     & 19h14m & 55.91 & +4.37 \\ \bottomrule
\end{NiceTabular}
\caption{The time costs and gains for different policy gradient algorithms.}
\label{tab:time}
\end{table*}

In this section, we analyze the efficiency and computational cost of our method, and compare it against other policy gradient algorithms in terms of training time and performance gains. We consider two representative reinforcement learning baselines: GRPO and DAPO~\citep{yu2025dapo}. The latter is a widely adopted variant of GRPO that incorporates several advanced techniques, including dynamic sampling and higher clipping ratio. As shown in Table~\ref{tab:time}, under the same data budget and training epochs, DAPO requires approximately $6$ additional hours of training time compared to GRPO, whereas our method incurs only a marginal increase of around $30$ minutes. More importantly, this modest computational overhead yields a substantial performance gain: our method improves accuracy by $4.37\%$ over GRPO, significantly surpassing the benefits provided by DAPO.

This efficiency is largely attributable to the nature of our perceptual supervision. The key difference of our method from GRPO lies in the anchor scoring process, where the model is asked to evaluate visual claims. Unlike standard rollouts, which require generating full reasoning traces, our method only requires a binary judgment, \ie, yes or no for each claim, essentially a single-token output. Consequently, this process is highly efficient. Moreover, our gradient update procedure remains identical to that of GRPO. These results suggest that our approach introduces minimal additional computational cost while achieving notably greater performance improvements.

\begin{table*}[t]
\footnotesize
\centering
\setlength{\tabcolsep}{1.6mm}
\setlength\heavyrulewidth{0.2ex}
\renewcommand{\arraystretch}{1.0}
\begin{NiceTabular}{@{}c|ccc@{}}
\toprule
\ Dataset & Total & Querying & Filtering \\ \midrule
\ ViRL39K & 38min & 34min    & 4min      \\ \bottomrule
\end{NiceTabular}
\caption{The time costs for generating visual claims.}
\label{tab:time_generate}
\end{table*}

\textcolor{blue}{Beyond training, In VAPO, we leverage LLMs, typically GPT to generate visual claims for assessing the model's perception capability. Here for completeness, we provide below an estimate of the computational time required to generate these visual claims, based on our usage records. As shown in Table~\ref{tab:time_generate}, the time is mainly spent on querying GPT and on the subsequent filtering process. The querying stage refers to the waiting time after submitting prompts, which also includes re-queries for generated claims that fail to meet the specified requirements. The filtering stage applies a set of rule-based criteria as described in Fig.~\ref{fig:claim_prompt} to remove claims that do not satisfy our standards, for example, those that are excessively long, contain uncertainty markers such as ``${\rm maybe}$'' or ``${\rm appears}$'', or violate the required balance between correct and incorrect claims. Besides, we also remove redundant claims that are semantically repetitive with embedding models, and limit the number of claims from various types, \eg, color, relational and OCR, through regular expressions. Any filtered-out examples are resubmitted until the target number of valid claims is reached. Overall, the entire process takes roughly half an hour, which is negligible compared with the training phase and remains highly efficient relative to methods such as DAPO even considering the overall pipeline costs. It is important to note that this claim generation process is a one-time procedure, and no further GPT querying is required for subsequent training runs.}

\subsection{B.10 \quad Limitation Analysis}

\textbf{Visual Claim Quality}. One factor influencing the effectiveness of \texttt{VAPO} is the quality of the generated visual claims. This includes whether the claims are strongly vision-dependent, clearly verifiable from the image, and correctly labeled. As such, the claim generation model, \ie, GPT-5 in our current setup, may become a limiting factor for overall performance. However, this also suggests that \texttt{VAPO} retains great potential for further improvement: leveraging higher-quality visual claims, either generated by more capable models or annotated by human experts, could potentially enhance the effectiveness and further improve the model's performance ceiling.

\textbf{Hyperparameter Sensitivity}. Our method introduces several new hyperparameters including the anchor number $K$, the late-emphasis weight $\beta$, and the perception reward weight $\gamma$, all of which require careful tuning to achieve optimal performance. Moreover, the optimal settings for these hyperparameters may vary across task types: vision-intensive tasks tend to benefit from larger values of $\beta$ and $\gamma$, whereas logic-heavy tasks such as mathematical reasoning often prefer smaller values. Designing an adaptive vision-anchored policy that dynamically adjusts these parameters based on task characteristics remains an important direction for future work.

\textbf{Single Image Setting}. In the current work, we focus exclusively on single-image tasks for simplicity. However, \texttt{VAPO} can be readily extended to multi-image or even video-based tasks by generating visual claims that reference multiple frames or views. Exploring the benefits of vision-anchored training in such settings, particularly in multi-image reasoning or temporal video understanding presents an interesting and promising direction for future research.

\subsection{\textcolor{blue}{B.11 \quad Claim and Label Quality}}

\textcolor{blue}{Upon obtaining the generated claims, to assess the quality and potential noise of both the visual claims and their corresponding labels, here we conduct a straightforward yet reliable human evaluation as an indicator for their effectiveness prior to training. Specifically, we randomly sample 250 examples from the training set and evaluate each generated claim to determine whether it is visual-dependent, i.e., grounded in the visual cues of the image rather than deducible solely through textual reasoning, and whether it is answerable, i.e., has a clear and unambiguous ground-truth answer. For the labels, we manually verify the accuracy against the corresponding question and visual input. Below, we report the recorded human evaluation statistics for the claims and labels.}

\begin{table*}[t]
\footnotesize
\centering
\setlength{\tabcolsep}{1.6mm}
\setlength\heavyrulewidth{0.2ex}
\renewcommand{\arraystretch}{1.0}
\begin{NiceTabular}{@{}c|ccc@{}}
\toprule
\ Human Evaluation & Visual-Dependent (\%) & Answerable (\%) & Accuracy (\%) \\ \midrule
\ Claims           & 98.44                 & 99.30           & \NA             \\
\ Labels           & \NA                     & \NA               & 96.78         \\ \bottomrule
\end{NiceTabular}
\caption{The human audit of the generated claim and label quality.}
\label{tab:human_quality}
\end{table*}

\begin{table*}[t]
\footnotesize
\centering
\setlength{\tabcolsep}{1.6mm}
\setlength\heavyrulewidth{0.2ex}
\renewcommand{\arraystretch}{1.0}
\begin{NiceTabular}{@{}c|ccc@{}}
\toprule
\ Model Evaluation & Visual-Dependent (\%) & Answerable (\%) & Accuracy (\%) \\ \midrule
\ Claims           & 97.29                 & 99.67           & -             \\
\ Labels           & -                     & -               & 95.85         \\ \bottomrule
\end{NiceTabular}
\caption{The model evaluation of the generated claim and label quality.}
\label{tab:model_quality}
\end{table*}

\textcolor{blue}{As shown in Table~\ref{tab:human_quality}, the generated claims exhibit high overall quality according to human evaluation, where the noise in the claims occupy only around $2\%$, \ie, claims that are not visual-dependent or not answerable, and the labeling error rate is below $4\%$. In addition, for comprehensiveness, here we also perform model evaluation, in which we sample 2000 examples and feed the generated claims and labels back to GPT and prompt it to act as a judge to assess noise and accuracy as defined above. The above results in Table~\ref{tab:model_quality} indicate that model and human evaluation of claim quality are closely aligned, suggesting that the generated claims and labels are reliable. As highlighted in our limitation analysis, further reducing the noise and error rates has the potential to yield additional performance gains and further enhance the effectiveness of our approach.}

\begin{table*}[b]
\footnotesize
\centering
\setlength{\tabcolsep}{1.6mm}
\setlength\heavyrulewidth{0.2ex}
\renewcommand{\arraystretch}{1.0}
\begin{NiceTabular}{@{}llllllll@{}}
\toprule
Model      & MathVerse & LogicVista & Geometry3k & MMMU & MMStar & HallBench & Avg. \\ \midrule
GRPO       & 44.2      & 44.4       & 44.5       & 53.6 & 55.2   & 51.2      & 49.0 \\
LLaVA-1.6  & 48.1      & 46.7       & 47.0       & 56.5 & 58.6   & 55.0      & 52.0 \\
Qwen2.5-VL & 48.5      & 47.0       & 47.4       & 56.7 & 58.9   & 55.2      & 52.3 \\
GPT-5      & 48.9      & 47.3       & 47.7       & 56.7 & 59.1   & 55.5      & 52.5 \\ \bottomrule
\end{NiceTabular}
\caption{The VAPO effectiveness given claims generated with other models.}
\label{tab:generators}
\end{table*}

\subsection{\textcolor{blue}{B.12 \quad Method Sensitivity to Other Generators}}
\textcolor{blue}{In our main experiments, we employ GPT-5 to generate visual claims due to its strong performance and cost-effectiveness. Here we further examine how using open-source, weaker models to generate visual claims influences VAPO's effectiveness. Specifically, we consider two typical VLMs: LLaVA-1.6-34B~\citep{liu2023visual} and Qwen2.5-VL-32B~\citep{bai2025qwen2}. For each model, we input the instructions directly as user prompts and apply a filtering mechanism similar to that used with GPT-5 to remove claims that do not meet our requirements. We train VAPO on a subset of 5000 examples randomly sampled from the full training set using the claims generated by these models, and evaluate results on the established benchmarks. As shown in Table~\ref{tab:generators}, despite being open-source models, Qwen2.5-VL achieves performance comparable to those obtained with GPT-5, yielding nearly identical average improvements over GRPO ($\pm 0.2\%$). Similarly, LLaVA-1.6, one of the pioneering VLMs, also attains nearly the full performance achieved with GPT-5 ($\pm 0.5\%$). These results indicate that VAPO does not have to rely on closed-source, state-of-the-art models to be effective; smaller open-source models are capable of delivering almost the same performance gains.}

\subsection{\textcolor{blue}{B.13 \quad VAPO on Other Models}}

\textcolor{blue}{Following prior work~\citep{huang2025vision, vl-rethinker}, we use Qwen2.5-VL as a standard base model to evaluate the effectiveness of our method, given its strong and representative reasoning capabilities. We posit that visual forgetting is primarily driven by two factors: 1) Training data. The existing training corpora (e.g., mathematics or text-intensive tasks) place heavy emphasis on textual patterns, leading models to develop a strong text bias; 2) Training objective. The widespread use of next-token prediction objective encourages models to prioritize autoregressive consistency, causing attention to visual inputs to become progressively diluted as the output sequence grows, as shown in Fig.~\ref{fig:forget} of the main paper. Therefore, unless these underlying factors are substantially altered, we expect that visual forgetting will likely generalize across different model architectures.}

\begin{table*}[t]
\footnotesize
\centering
\setlength{\tabcolsep}{1.6mm}
\setlength\heavyrulewidth{0.2ex}
\renewcommand{\arraystretch}{1.0}
\begin{NiceTabular}{@{}lccccccc@{}}
\toprule
\multicolumn{1}{c}{Model} & MathVerse     & LogicVista    & Geometry3k    & MMMU          & MMStar        & HallBench     & Avg.          \\ \midrule
Qwen3-VL                  & 58.8          & 56.7          & 59.3          & 68.4          & 67.3          & 59.2          & 61.8          \\
\quad + GRPO                    & 63.5          & 60.1          & 64.4          & 72.7          & 68.3          & 60.4          & 64.9          \\
\quad + VAPO                    & \textbf{67.6} & \textbf{65.8} & \textbf{68.7} & \textbf{74.2} & \textbf{72.5} & \textbf{63.9} & \textbf{68.8} \\ \bottomrule
\end{NiceTabular}
\caption{The results of VAPO on other base models.}
\label{tab:qwen3}
\end{table*}

\textcolor{blue}{To provide a preliminary validation of this assumption, here we additionally experiment with Qwen3-VL-8B~\citep{yang2025qwen3}, a more recent and more capable reasoning model whose architecture differ substantially from those of Qwen2.5-VL. We apply VAPO to Qwen3-VL to assess whether our method remains effective under different scenarios. As shown in Table~\ref{tab:qwen3}, our method continues to provide substantial improvements over GRPO when applied to Qwen3-VL, yielding an average gain of $3.9\%$ ($64.9\% \rightarrow 68.8\%$). This suggests that, beyond Qwen2.5-VL, other reasoning models may also suffer from pronounced visual forgetting, and that our method is broadly applicable and effective across different model architectures.}

\subsection{\textcolor{blue}{B.14 \quad Evaluation on More Tasks}}

\begin{table*}[t]
\footnotesize
\centering
\setlength{\tabcolsep}{1.6mm}
\setlength\heavyrulewidth{0.2ex}
\renewcommand{\arraystretch}{1.0}
\begin{NiceTabular}{@{}lccc@{}}
\toprule
\multicolumn{1}{c}{Model} & V*            & RealWorldQA   & Avg.          \\ \midrule
Qwen2.5-VL                & 71.4          & 64.3          & 67.9          \\
R1-OneVision              & 73.5          & 65.7          & 69.6          \\
VLAA-Thinker              & 74.8          & 66.3          & 70.6          \\
Vision-R1                 & 74.1          & 66.8          & 70.5          \\
VAPO-Thinker              & \textbf{77.9} & \textbf{69.4} & \textbf{73.7} \\ \bottomrule
\end{NiceTabular}
\caption{The results of VAPO on more evaluation tasks.}
\label{tab:more_tasks}
\end{table*}

\textcolor{blue}{During training, our visual claims are designed to span multiple dimensions of perceptual capability, including fine-grained visual details, \eg, ``the whiteboard says No Parking'', and spatial reasoning, \eg, ``the bus is to the left of the motorcycle''. Consequently, our perception training enhances performance on tasks requiring these capabilities. For completeness, here we further evaluate our model on two popular benchmarks, V*~\citep{wu2024v} and RealWorldQA, which target fine-grained perception and spatial reasoning, respectively. As shown in Table~\ref{tab:more_tasks}, our model achieves an average improvement of $3.1\%$ ($70.6\% \rightarrow 73.7\%$) over the previous best results on both fine-grained and spatial reasoning benchmarks. This empirically demonstrates that VAPO generalizes well to a broad range of perception-intensive tasks.}

\subsection{\textcolor{blue}{B.15 \quad VAPO with Self-Supervised Anchors}}

\begin{table*}[b]
\footnotesize
\centering
\setlength{\tabcolsep}{1.6mm}
\setlength\heavyrulewidth{0.2ex}
\renewcommand{\arraystretch}{1.0}
\begin{NiceTabular}{@{}lccccccc@{}}
\toprule
\multicolumn{1}{c}{Model} & MathVerse & LogicVista & Geometry3k & MMMU & MMStar & HallBench & Avg. \\ \midrule
Base Model                & 40.7      & 42.6       & 38.5       & 52.7 & 54.9   & 50.0      & 46.6 \\
GRPO                      & 44.2      & 44.4       & 44.5       & 53.6 & 55.2   & 51.2      & 49.0 \\
VAPO (Self-Supervised)    & 48.1      & 46.5       & 48.3       & 55.9 & 57.9   & 55.1      & 51.9 \\
VAPO (GPT)                & 48.9      & 47.3       & 47.7       & 56.7 & 59.1   & 55.5      & 52.5 \\ \bottomrule
\end{NiceTabular}
\caption{The results of VAPO with self-supervised anchors.}
\label{tab:self-supervised}
\end{table*}

\textcolor{blue}{The core idea of VAPO is to utilize anchors to assess model's perception capability during reasoning process and use curated rewards to encourage stronger reliance on visual inputs. The specific form of the anchor (i.e., the mechanism used to measure perception), is not essentially limited to binary textual claims. We choose textual claims mainly because 1) efficiency: As discussed in the efficiency analysis, current pipeline introduces only minimal training overhead, offering an excellent cost-performance trade-off; 2) simplicity, unlike reconstruction-based methods which may require additional learnable decoders or more complex architectures, textual claims allow us to probe perception capability with only a single-token output; 3) effectiveness: Our experiments also demonstrate that textual claims significantly improve model accuracy, surpassing previous strong baselines.}

\textcolor{blue}{When designing the visual anchors, we also considered reconstruction-based anchors, where the model would be required to reconstruct visual features or undergo linear probing to assess its perceptual capability. However, as mentioned above, these approaches introduce additional learnable parameters, thereby increasing computational and training complexity. }

\textcolor{blue}{Another promising self-supervised direction is to adopt a bootstrapping approach, in which the model generates its own visual claims and subsequently uses them as training signals, thereby eliminating the need for external models. Here  we explore whether this more efficient training strategy for VAPO remains effective. Specifically, instead of relying on GPT to produce visual claims, we allow the model being trained to generate its own visual claims during the rollout step using the provided instructions, and then use these self-generated claims to assess its perceptual capability and update the model accordingly. This procedure is self-supervised and iterative: the model’s generated claims help improve its visual capability, which in turn further refines the quality of subsequent claims. We use Qwen2.5-VL-7B as the base model and train on a 5000 example subset of the full training set. As shown in Table~\ref{tab:self-supervised}, this self-supervised approach retains the vast majority of the effectiveness observed in the default (GPT) setting, with only a modest average difference of $0.7\%$ ($46.5\% \rightarrow 47.3\%$). This indicates that, even without relying on any external or stronger models, VAPO can still achieve highly promising performance.}

\subsection{\textcolor{blue}{B.16 \quad Single-cut Recoverable Ratio}}

\begin{table*}[t]
\footnotesize
\centering
\setlength{\tabcolsep}{1.6mm}
\setlength\heavyrulewidth{0.2ex}
\renewcommand{\arraystretch}{1.0}
\begin{NiceTabular}{@{}c|ccccc@{}}
\toprule
Cut Position (\%)      & 20  & 40  & 60  & 80  & Agg. \\ \midrule
Recoverable Ratio (\%) & 4.5 & 5.1 & 8.8 & 7.6 & 15.4 \\ \bottomrule
\end{NiceTabular}
\caption{The single-cut recoverable ratio by early decision.}
\label{tab:single-cut}
\end{table*}

\textcolor{blue}{Our current definition of the recoverable ratio uses a relatively loose metric intended to estimate the proportion of error cases influenced by the reasoning process, serving as an oracle-style indicator of the potential for our proposed method. Here, we additionally report recoverable ratios at specific single-cut positions. In particular, we provide the recoverable ratios measured at four relative reasoning depths ($20\%$, $40\%$, $60\%$, and $80\%$ of the reasoning length) as well as their aggregated result. As shown in Table~\ref{tab:single-cut}, even when considering only a single reasoning position, the model’s recoverable ratio remains substantial, reaching nearly $10\%$ at the $60\%$ and $80\%$ positions. This indicates considerable headroom for improving the accuracy of current models and helps explain why excessively long reasoning can degrade performance, as observed in Fig.~\ref{fig:perception}. Moreover, we observe that later positions exhibit higher recoverable ratios, suggesting that the latter stages of reasoning have a particularly strong impact on the model’s reliance on visual information, consistent with the visual forgetting phenomenon shown in Fig.~\ref{fig:forget}.}

\subsection{\textcolor{blue}{B.17 \quad Data Deduplication Check}}

\begin{table*}[b]
\footnotesize
\centering
\setlength{\tabcolsep}{1.6mm}
\setlength\heavyrulewidth{0.2ex}
\renewcommand{\arraystretch}{1.0}
\begin{NiceTabular}{@{}c|ccc@{}}
\toprule
Deduplication Check & Mathematical & General & Total \\ \midrule
Overlap Ratio (\%)  & 0            & 0       & 0     \\ \bottomrule
\end{NiceTabular}
\caption{The overlap ratio between training and evaluation set.}
\label{tab:overlap}
\end{table*}

\textcolor{blue}{In our training setup, we follow prior work~\citep{vl-rethinker, wang2025perception} by training on ViRL39K and evaluating on both mathematical benchmarks (e.g., MathVista) and general-purpose benchmarks (e.g., MMStar). Here we perform data deduplication checks between the training set and the evaluation benchmarks. Specifically, we compare image hash values across datasets and compute the number of overlapping instances. As shown in Table~\ref{tab:overlap}, both the mathematical benchmarks (6 in total) and the general-purpose benchmarks (4 in total) exhibit $0\%$ overlap with the training dataset based on image-hash comparison. This indicates that no duplicate samples exist between the training and evaluation sets, and thus there is no risk of data contamination.}

\subsection{\textcolor{blue}{B.18 \quad Effect of Accuracy Gate}}

\textcolor{blue}{In training, we incorporate an accuracy gate to prevent reward hacking, where the model might pursue perception rewards at the expense of task accuracy by producing trivially short reasoning. Although the accuracy gate is triggered more frequently on easy problems, resulting in a larger portion of trajectories receiving perception rewards, these rewards are ultimately normalized when converted into advantages. This normalization step ensures that the generally higher rewards obtained from easy examples do not disproportionately influence the final advantage. In contrast, for hard problems, normalization scales up the advantages of correct trajectories, encouraging the model to pay additional attention to these challenging cases. Therefore, the accuracy gate does not bias training toward simpler examples.}

\begin{table*}[t]
\footnotesize
\centering
\setlength{\tabcolsep}{1.6mm}
\setlength\heavyrulewidth{0.2ex}
\renewcommand{\arraystretch}{1.0}
\begin{NiceTabular}{@{}c|ccccc@{}}
\toprule
Training Steps         & 0    & 25   & 50   & 75   & 100  \\ \midrule
Activation Rate (Easy) & 0.53 & 0.65 & 0.72 & 0.76 & 0.79 \\
Activation Rate (Hard) & 0.21 & 0.44 & 0.56 & 0.61 & 0.64 \\ \bottomrule
\end{NiceTabular}
\caption{The accuracy gate activation gate across easy and hard examples.}
\label{tab:activation}
\end{table*}

\textcolor{blue}{To further analyze how the accuracy gate influences the training dynamics of easy and hard problems, here we report the accuracy gate activation rates over the course of training for both categories. We classify problems as easy or hard based on the model’s initial accuracy gate activation rate ($>0.4$ for easy and $<0.4$ for hard) and track how these activation rates evolve at different training steps. We conduct training on a $5000$ example subset of the full training set. As shown in Table~\ref{tab:activation}, the activation rate increases substantially for both easy and hard questions, and the gap between the two progressively narrows. This further indicates that the model does not over or under-optimize either category during training, suggesting balanced learning dynamics across problem difficulties.}

\begin{table*}[t]
\footnotesize
\centering
\setlength{\tabcolsep}{1.6mm}
\setlength\heavyrulewidth{0.2ex}
\renewcommand{\arraystretch}{1.0}
\begin{NiceTabular}{@{}lccccccc@{}}
\toprule
\multicolumn{1}{c}{Method} & MathVerse     & LogicVista    & Geometry3k    & MMMU          & MMStar        & HallBench     & Avg.          \\ \midrule
GRPO                       & 44.2          & 44.4          & 44.5          & 53.6          & 55.2          & 51.2          & 49.0          \\
VAPO (strict gate)         & \textbf{48.9} & 47.3          & \textbf{47.7} & \textbf{56.7} & 59.1          & \textbf{55.5} & \textbf{52.5} \\
VAPO (relaxed gate)        & 48.5          & \textbf{47.4} & 47.2          & 56.3          & \textbf{59.4} & 55.0          & 52.3          \\ \bottomrule
\end{NiceTabular}
\caption{The results of the curriculum gating strategy.}
\label{tab:relaxed}
\end{table*}

\textcolor{blue}{Moreover, to investigate whether a relaxed accuracy gate can further improve performance, we experiment with a curriculum gating strategy. Specifically, unlike our default setting, which applies a strict binary gate, we gradually loosen this constraint over the course of training. That is, we progressively assign an increasing proportion of the perception reward to incorrect trajectories and eventually remove the accuracy gate entirely by granting full perception rewards. As before, we conduct this study using a 5000 example subset of the full training set. As shown in Table~\ref{tab:relaxed}, using a relaxed accuracy gate does not yield noticeable performance improvements. This aligns with our earlier analysis that the default accuracy gate does not under-optimize hard examples, and that relaxing the gate introduces a greater risk of reward hacking.}

\subsection{\textcolor{blue}{B.19 \quad Effect of Anchor Placement Strategy}}

\textcolor{blue}{Our current anchor placement strategy resembles the early decision scheme, where anchors are inserted at semantic boundaries (e.g., commas, periods, line breaks) randomly throughout the entire reasoning trajectory. In designing this component, we also considered several alternative placement strategies: 1) uniform, where anchors are inserted at fixed intervals rather than randomly; 2) back-loaded, where anchors are inserted only in the latter $60\%$ of the reasoning process, acting as a targeted mechanism to address severe visual forgetting that tends to occur toward the end of long trajectories; 3) front-loaded, where anchors are inserted only within the first $60\%$ of the trajectory, serving as a baseline comparison. We ensure that all three variants insert the same number of anchors ($K=20$). We train each variant on a $5000$ example subset of the full training set, and we report the results of these placement strategies.}

\begin{table*}[t]
\footnotesize
\centering
\setlength{\tabcolsep}{1.6mm}
\setlength\heavyrulewidth{0.2ex}
\renewcommand{\arraystretch}{1.0}
\begin{NiceTabular}{@{}cccccccc@{}}
\toprule
Position     & MathVerse     & LogicVista    & Geometry3k    & MMMU          & MMStar        & HallBench     & Avg.          \\ \midrule
GRPO         & 44.2          & 44.4          & 44.5          & 53.6          & 55.2          & 51.2          & 49.0          \\
Uniform      & 48.4          & 47.1          & 47.5          & 56.1          & 58.4          & 55.0          & 52.1          \\
Front-loaded & 48.1          & 46.7          & 47.0          & 55.9          & 58.5          & 54.4          & 51.8          \\
Back-loaded  & 48.7          & 46.9          & 47.7          & 56.2          & 59.0          & \textbf{55.8} & 52.4          \\
Random       & \textbf{48.9} & \textbf{47.3} & \textbf{47.7} & \textbf{56.7} & \textbf{59.1} & 55.5          & \textbf{52.5} \\ \bottomrule
\end{NiceTabular}
\caption{The effect of varying anchor placement strategies.}
\label{tab:position}
\end{table*}

\textcolor{blue}{As shown in Table~\ref{tab:position}, the simple naive (random) strategy performs better than other placement variants. We attribute this to several factors: 1) uniform insertion, which places anchors at fixed intervals, is more susceptible to reward hacking, as the model may learn to exhibit high visual dependence only at predetermined positions; 2) front-loaded insertion performs noticeably worse than both random and back-loaded insertion, indicating that later positions in the reasoning trajectory are more critical for improving perceptual grounding; 3) In addition, although back-loaded insertion is slightly weaker than random insertion on average, it achieves comparable or even superior performance on MMStar and HallusionBench. This suggests that vision-heavy benchmarks may benefit from a larger number of anchors or greater weighting at later reasoning steps, which aligns with our late-emphasis design.}

\subsection{\textcolor{blue}{B.20 \quad Comparison with More Baselines}}

\begin{table*}[t]
\footnotesize
\centering
\setlength{\tabcolsep}{1.6mm}
\setlength\heavyrulewidth{0.2ex}
\renewcommand{\arraystretch}{1.0}
\begin{NiceTabular}{@{}lccccccc@{}}
\toprule
\multicolumn{1}{c}{Method} & MathVerse     & LogicVista    & Geometry3k    & MMMU          & MMStar        & HallBench     & Avg.          \\ \midrule
Base Model                 & 40.7          & 42.6          & 38.5          & 52.7          & 54.9          & 50.0          & 46.6          \\
MM-Eureka                  & 50.3          & 48.5          & 50.2          & 55.7          & 60.4          & 54.8          & 53.3          \\
PAPO                       & 50.4          & 47.3          & 49.7          & \textbf{60.8} & 61.8          & 55.6          & 54.2          \\
VL-Rethinker               & \textbf{54.2} & 47.2          & 49.6          & 57.9          & 61.8          & 55.4          & 54.4          \\
VAPO                       & 53.3          & \textbf{50.9} & \textbf{51.3} & 60.2          & \textbf{63.0} & \textbf{57.4} & \textbf{56.0} \\ \bottomrule
\end{NiceTabular}
\caption{The comparison between VAPO with more baselines.}
\label{tab:more_baselines}
\end{table*}

\textcolor{blue}{Here, for completeness, we compare our approach with more baselines, including PAPO~\citep{wang2025perception}, MM-Eureka~\citep{meng2025mm} and VL-Rethinker~\citep{vl-rethinker}. We use the publicly released 7B checkpoints from their repositories. For PAPO, we reproduce the results using greedy decoding, as their paper reports performance using the avg@8 accuracy metric, which is not directly comparable to the metrics adopted in prior work~\citep{vl-rethinker, huang2025vision}. As shown in Table~\ref{tab:more_baselines}, VAPO achieves an average improvement of $1.6\%$ over the previous best results ($54.4\% \rightarrow 56.0\%$), further demonstrating the effectiveness of our approach.}

\subsection{\textcolor{blue}{B.21 \quad Effect of VAPO on Long-thought Examples}}

\begin{table*}[b]
\footnotesize
\centering
\setlength{\tabcolsep}{1.6mm}
\setlength\heavyrulewidth{0.2ex}
\renewcommand{\arraystretch}{1.0}
\begin{NiceTabular}{@{}c|ccccc@{}}
\toprule
Length & 1    & 5    & 10   & 15   & 20   \\ \midrule
Short  & 53.7 & 56.0 & 57.5 & 58.4 & 58.5 \\
Long   & 47.3 & 49.4 & 50.9 & 51.5 & 51.8 \\
Total  & 51.5 & 53.8 & 55.1 & 55.8 & 55.9 \\ \bottomrule
\end{NiceTabular}
\caption{The results of VAPO on short and long-thought examples.}
\label{tab:long}
\end{table*}

\textcolor{blue}{Our ablation study in Fig.~\ref{fig:ablation} shows that increasing $K$, \ie, using more anchors, generally yields positive gains across examples. To further examine whether longer-thought examples benefit disproportionately from larger $K$, we partition benchmark examples into two groups based on the output length of our VAPO-Thinker-7B model: short-thought ($<300$ tokens) and long-thought ($>300$ tokens). We then evaluate models trained with different values of $K$ and analyze how performance trends vary across these two categories. As shown in Table~\ref{tab:long}, increasing $K$ leads to stable performance improvements for both short- and long-thought examples, and the magnitude of these gains remains largely consistent across the two groups. This trend is also aligned with the ablation results reported in Fig.~\ref{fig:ablation}.}

\subsection{\textcolor{blue}{B.22 \quad Reasoning Length Distribution}}

\begin{table*}[t]
\footnotesize
\centering
\setlength{\tabcolsep}{1.6mm}
\setlength\heavyrulewidth{0.2ex}
\renewcommand{\arraystretch}{1.0}
\begin{NiceTabular}{@{}lccccccc@{}}
\toprule
\multicolumn{1}{c}{Models} & MathVerse & LogicVista & Geometry3k & MMMU & MMStar & HallBench & Avg. \\ \midrule
R1-OneVision               & 299       & 308        & 287        & 263  & 232    & 201       & 265  \\
Vision-R1                  & 337       & 293        & 341        & 283  & 239    & 251       & 291  \\
VAPO-Thinker               & 315       & 287        & 334        & 275  & 254    & 238       & 284  \\ \bottomrule
\end{NiceTabular}
\caption{The average reasoning length of VAPO.}
\label{tab:length}
\end{table*}

\textcolor{blue}{The goal of VAPO is to enhance the effectiveness of reasoning by improving the model’s perceptual capability, rather than to alter the length of its reasoning traces. Accordingly, our pipeline does not incorporate any explicit length regularization. Here we report the average reasoning length of our models compared with other baselines across the established benchmarks. As shown in Table~\ref{tab:length}, the average output length of our model does not differ noticeably from that of other baselines. This rules out the possibility that our performance gains stem merely from longer reasoning traces, and instead indicates that VAPO improves the effectiveness of the reasoning process itself.}

\subsection{\textcolor{blue}{B.23 \quad VAPO on Multi-image Task}}

\textcolor{blue}{Our current visual claims are primarily designed for single-image inputs, which is sufficient to validate the core motivation of VAPO, namely improving the model’s reliance on visual information and mitigating visual forgetting. Here we further conduct a small-scale pilot study on multi-image tasks. Specifically, we train VAPO exclusively on the multi-image subset of ViRL39K, which contains approximately 2,500 examples. For claim generation, we consider two types of claims. The first type is single-image claims, which mirror our original setup and are generated with respect to only one image in the multi-image set. The second type is cross-image claims, which capture relationships or distinctions across images. For the former, we provide GPT with only one image at a time. For the latter, we present all images simultaneously and extend the original instructions with an additional requirement that the generated claim must reference information across images (e.g., ``the two images depict the same building''). We ensure that each training example contains a balanced number of claims from both categories. The training procedure remains identical to our main pipeline, and GRPO is used as the baseline for comparison. We evaluate the trained models on two representative multi-image benchmarks, including BLINK~\citep{fu2024blink} and MuirBench~\citep{wang2024muirbench}.}

\begin{table*}[t]
\footnotesize
\centering
\setlength{\tabcolsep}{1.6mm}
\setlength\heavyrulewidth{0.2ex}
\renewcommand{\arraystretch}{1.0}
\begin{NiceTabular}{@{}cccc@{}}
\toprule
Method     & BLINK & MuirBench & Avg. \\ \midrule
Base model & 55.3  & 58.1      & 56.7 \\
GRPO       & 57.0  & 60.4      & 58.7 \\
VAPO       & 60.1  & 62.2      & 61.2 \\ \bottomrule
\end{NiceTabular}
\caption{The results of VAPO on multi-image tasks.}
\label{tab:multi-image}
\end{table*}

\textcolor{blue}{As shown in Table~\ref{tab:multi-image}, despite being trained on a limited amount of data, our method still achieves an average improvement of $2.5\%$ over GRPO on multi-image tasks ($58.7\% \rightarrow 61.2\%$). This demonstrates that our approach is not restricted to the single-image setting and can generalize to multiple images as well. We leave the evaluation of video-based tasks as promising future work.}

\section{C. \quad Broader Societal Impacts}

Our work carries several positive societal implications. Our findings reveal that existing VLMs often produce reasoning that is not visually grounded, which can degrade performance and reliability. Such ungrounded reasoning may introduce misleading or hallucinated content, thereby posing risks to model safety and trustworthiness. Moreover, since these reasoning traces often rely heavily on statistical patterns learned from training data, they may inadvertently expose sensitive information, raising concerns about data privacy and potential leakage. In contrast, the proposed \texttt{VAPO} framework steers the model’s reasoning process to remain anchored in visual evidence. This grounding mechanism helps ensure that model outputs are more faithful to the input image and less likely to include irrelevant or speculative content, thus promoting both factual accuracy and privacy preservation. At present, we have not identified negative societal impacts associated with this work. However, due to external factors such as the availability of datasets and baseline implementations, further assessment may be necessary in the future.

\section{D. \quad Supplementary Results}
\subsection{D.1 \quad Full Numerical Main Results}
Due to space limitations in the main text, some baseline results are omitted. Here, we provide the full numerical results of our method and all baselines across the ten established benchmarks in Table~\ref{tab:main}, corresponding to Table~\ref{tab:math} and Table~\ref{tab:general}.

\subsection{D.2 \quad Numerical Results of $K$}

Here we report the numerical results of the ablation study on the anchor number $K$ in Table~\ref{tab:K}, corresponding to the visualization in Fig.~\ref{fig:ablation}~(A).

\subsection{D.3 \quad Numerical Results of $\beta$}

Similarly, we present the numerical results of the ablation study on the impact of $\beta$ in Table~\ref{tab:beta}, corresponding to Fig.~\ref{fig:ablation}~(B).

\subsection{D.4 \quad Full results of Augmented Baselines}

Here we provide the full numerical results of our method compared with baselines augmented with inference-level remedies, \ie, visual replay and focus prompt, in Table~\ref{tab:full_inference}, corresponding to Table~\ref{tab:inference} in the main paper.

\clearpage
\begin{table*}[t]
\footnotesize
\centering
\setlength{\tabcolsep}{1.0mm}
\setlength\heavyrulewidth{0.2ex}
\renewcommand{\arraystretch}{1.4}
\begin{adjustbox}{width=1.0\textwidth,center}
\begin{NiceTabular}{@{}lccccccccccc@{}}
\toprule
\multicolumn{1}{c}{Models}   & MathVerse            & MathVista            & MathVision           & LogicVista           & WeMath               & Geo3k                & MMMU                 & MMStar               & HallBench            & MMVet                & Avg.                  \\ \midrule
\textit{Close-source Models} & \multicolumn{1}{l}{} & \multicolumn{1}{l}{} & \multicolumn{1}{l}{} & \multicolumn{1}{l}{} & \multicolumn{1}{l}{} & \multicolumn{1}{l}{} & \multicolumn{1}{l}{} & \multicolumn{1}{l}{} & \multicolumn{1}{l}{} & \multicolumn{1}{l}{} & \multicolumn{1}{l}{} \\
GPT-5-Thinking               & 81.2                 & 81.9                 & 72.0                 & 70.0                 & 71.1                 & 79.9                 & 81.8                 & 75.7                 & 65.2                 & 77.6                 & 75.6                 \\
Gemini-2.5-Pro               & 76.9                 & 80.9                 & 69.1                 & 73.8                 & 78.0                 & 77.2                 & 74.7                 & 73.6                 & 64.1                 & 83.3                 & 75.2                 \\ \midrule
\textit{Open-source Models}  & \multicolumn{1}{l}{} & \multicolumn{1}{l}{} & \multicolumn{1}{l}{} & \multicolumn{1}{l}{} & \multicolumn{1}{l}{} & \multicolumn{1}{l}{} & \multicolumn{1}{l}{} & \multicolumn{1}{l}{} & \multicolumn{1}{l}{} & \multicolumn{1}{l}{} & \multicolumn{1}{l}{} \\
Qwen2.5-VL-7B                & 40.7                 & 62.3                 & 23.2                 & 42.6                 & 33.1                 & 38.5                 & 52.7                 & 54.9                 & 50.0                 & 64.8                 & 46.3                 \\
InternVL2.5-8B               & 34.5        & 68.2        & 25.6        & 38.3        & 38.6        & 44.8        & 56.2        & 63.2                 & 49.0                 & 62.8                 & 48.1                 \\
R1-OneVision-7B              & 46.4                 & 64.1                 & 29.9                 & 45.6                 & \textbf{44.6}        & 46.1                 & 54.3                 & 54.1                 & 52.5                 & 65.2                 & 50.3                 \\
VLAA-Thinker-7B              & 48.2                 & 68.0                 & 26.4                 & 48.5                 & 41.5                 & 50.6                 & 59.1                 & 49.7                 & 54.7                 & 70.0                 & 51.7                 \\
Vision-R1-7B                 & 52.4                 & 73.5                 & 28.2                 & 49.7                 & 41.6                 & 49.0                 & 57.6                 & 61.4                 & 49.5                 & 71.1                 & 53.4                 \\ \midrule
\textit{Our Models}          & \multicolumn{1}{l}{} & \multicolumn{1}{l}{} & \multicolumn{1}{l}{} & \multicolumn{1}{l}{} & \multicolumn{1}{l}{} & \multicolumn{1}{l}{} & \multicolumn{1}{l}{} & \multicolumn{1}{l}{} & \multicolumn{1}{l}{} & \multicolumn{1}{l}{} & \multicolumn{1}{l}{} \\
\texttt{VAPO-Thinker-3B}              & 35.8                 & 67.1                 & 23.9                 & 39.7                 & 35.4                 & 44.2                 & 55.6                 & 59.4                 & 49.5                 & 64.6                 & 47.5                 \\
\rowcolor[HTML]{EDEDED} \texttt{VAPO-Thinker-7B}              & \textbf{53.3}        & \textbf{75.6}        & \textbf{31.9}        & \textbf{50.9}        & 43.6                 & \textbf{51.3}        & \textbf{60.2}        & \textbf{63.0}        & \textbf{57.4}        & \textbf{71.9}        & \textbf{55.9}        \\ \bottomrule
\end{NiceTabular}
\end{adjustbox}
\caption{The full numerical results of main experiments, corresponding to Table~\ref{tab:math} and Table~\ref{tab:general}.}
\label{tab:main}
\end{table*}

\clearpage
\begin{table*}[t]
\footnotesize
\centering
\setlength{\tabcolsep}{1.0mm}
\setlength\heavyrulewidth{0.2ex}
\renewcommand{\arraystretch}{1.4}
\begin{adjustbox}{width=1.0\textwidth,center}
\begin{tabular}{@{}cccccccccccc@{}}
\toprule
$K$  & MathVerse      & MathVista      & MathVision     & LogicVista     & WeMath         & Geo3k          & MMMU           & MMStar         & HallBench      & MMVet          & Avg.            \\ \midrule
0  & 48.2          & 70.6          & 26.1          & 45.5          & 39.1          & 47.3          & 56.6          & 58.9          & 53.2          & 69.9          & 51.5          \\
5  & 50.9          & 73.1          & 28.4          & 47.8          & 41.2          & 50.5          & 58.4          & 61.3          & 55.6          & 70.4          & 53.8          \\
10 & 52.8          & 74.8          & 30.7          & 49.5          & 42.5          & 51.2          & 59.9          & 62.2          & 56.7          & 71.2          & 55.1          \\
15 & 53.2          & \textbf{75.8} & 31.5          & 50.4          & \textbf{43.9} & 51.0          & \textbf{60.4} & 62.7          & 57.0          & 71.6          & 55.8          \\
20 & \textbf{53.3} & 75.6          & \textbf{31.9} & \textbf{50.9} & 43.6          & \textbf{51.3} & 60.2          & \textbf{63.0} & \textbf{57.4} & \textbf{71.9} & \textbf{55.9} \\ \bottomrule
\end{tabular}
\end{adjustbox}
\caption{The numerical results of ablation study of $K$, corresponding to Fig.~\ref{fig:ablation}~(A).}
\label{tab:K}
\end{table*}

\clearpage
\begin{table*}[t]
\footnotesize
\centering
\setlength{\tabcolsep}{1.0mm}
\setlength\heavyrulewidth{0.2ex}
\renewcommand{\arraystretch}{1.4}
\begin{adjustbox}{width=1.0\textwidth,center}
\begin{tabular}{@{}cccccccccccc@{}}
\toprule
$\beta$   & MathVerse      & MathVista      & MathVision     & LogicVista     & WeMath         & Geo3k          & MMMU           & MMStar         & HallBench      & MMVet          & Avg            \\ \midrule
0.0 & 51.7          & 73.8          & 28.4          & 48.6          & 42.0          & 49.9          & 58.9          & 60.2          & 53.8          & 70.3          & 53.8          \\
0.5 & 52.7          & 74.8          & 30.2          & 49.6          & 42.8          & 50.7          & 59.5          & 62.0          & 55.4          & 71.3          & 54.9          \\
1.0 & 53.1          & \textbf{75.7} & 31.5          & \textbf{51.3} & 43.4          & 51.4          & 60.0          & 62.5          & 56.8          & \textbf{72.0} & 55.8          \\
1.5 & \textbf{53.3} & 75.6 & \textbf{31.9} & 50.9          & \textbf{43.6} & 51.3          & \textbf{60.2} & 63.0          & 57.4 & 71.9          & \textbf{55.9} \\
2.0 & 53.0 & 75.5          & 31.1 & 50.5 & 43.1          & \textbf{51.5} & 60.0          & \textbf{63.3} & \textbf{56.6} & 71.6 & 55.6 \\
2.5 & 51.8          & 74.8          & 30.4          & 50.0          & 42.6          & 50.5          & 59.8          & 62.2          & 55.3          & 70.2          & 54.7          \\ \bottomrule
\end{tabular}
\end{adjustbox}
\caption{The numerical results of ablation study of $\beta$, corresponding to Fig.~\ref{fig:ablation}~(B).}
\label{tab:beta}
\end{table*}

\clearpage
\begin{table*}[t]
\footnotesize
\centering
\setlength{\tabcolsep}{1.0mm}
\setlength\heavyrulewidth{0.2ex}
\renewcommand{\arraystretch}{1.4}
\begin{adjustbox}{width=1.0\textwidth,center}
\begin{NiceTabular}{@{}lccccccccccc@{}}
\toprule
\multicolumn{1}{c}{\textbf{Models}} & MathVerse     & MathVista     & MathVision    & LogicVista    & WeMath        & Geo3k         & MMMU          & MMStar        & HallBench     & MMVet         & Avg           \\ \midrule
VLAA-Thinker-7B                     & 48.2          & 68.0          & 26.4          & 48.5          & 41.5          & 50.6          & 59.1          & 49.7          & 54.7          & 70.0          & 51.7          \\
\quad + Focus Prompt                      & 48.8          & 68.4          & 27.3          & 49.2          & 41.8          & 50.7          & 59.7          & 51.1          & 55.2          & 70.7          & 52.3          \\
\quad + Visual Replay                     & 49.8          & 69.8 & 27.9          & 49.9 & 42.3          & 51.1          & 60.5          & 52.9          & 56.2          & 71.7 & 53.2          \\
Vision-R1-7B                        & 52.4 & 73.5 & 28.2 & 49.7          & 41.6 & 49.0          & 57.6 & 61.4          & 49.5 & 71.1          & 53.4 \\
\quad + Focus Prompt                      & 52.8 & 73.8          & 29.3 & 50.2 & 42.1          & 49.7 & 58.7          & 61.8 & 50.5 & 71.4 & 54.0 \\
\quad + Visual Replay                     & \textbf{53.4} & 75.2          & 29.9          & 50.7          & 42.5          & 50.5          & 59.4          & 62.1          & 51.8          & 71.9          & 54.7          \\
\rowcolor[HTML]{EDEDED} \texttt{VAPO-Thinker-7B}                     & 53.3          & \textbf{75.6} & \textbf{31.9} & \textbf{50.9} & \textbf{43.6} & \textbf{51.3} & \textbf{60.2} & \textbf{63.0} & \textbf{57.4} & \textbf{71.9} & \textbf{55.9} \\ \bottomrule
\end{NiceTabular}
\end{adjustbox}
\caption{The full results of baselines augmented with test-time remedies, corresponding to Table~\ref{tab:inference}.}
\label{tab:full_inference}
\end{table*}

\end{document}